%% file: main.tex
\documentclass{arxivtmpl}

\usepackage[T1]{fontenc}

\usepackage{amsmath,amsfonts,bm}
\usepackage{xspace}

\newcommand{\method}{NFWM\xspace}

\usepackage{amssymb,amsthm,mathtools}
\usepackage{booktabs}
\usepackage{multirow}
\usepackage{float}
\usepackage{graphicx}
\usepackage{caption}

\usepackage{wrapfig}
\usepackage{placeins}
\usepackage{needspace}
\usepackage{microtype}
\usepackage{xcolor}
\usepackage{hyperref}
\usepackage{url}
\usepackage{fontawesome5}

\definecolor{titlered}{HTML}{A90533}
\renewcommand{\titlefont}{\normalfont\bfseries\fontsize{21}{26}\selectfont\color{titlered}}
\renewcommand\Affilfont{\sffamily\mdseries}

\title{Normalizing Flows are Capable World Models}

\author[1,*]{Zijia Li}
\author[1,*]{Yong-Hyun Park}
\author[2]{Shusheng Yang}
\author[1]{Jiatao Gu}
\affil[1]{University of Pennsylvania}
\affil[2]{New York University}
\authornote{\textsuperscript{*}Equal contribution.\\[2pt]
  {\hypersetup{urlcolor=black}%
  \href{https://peter-lizijia.github.io/NFWM-Project-Page/}{\faGlobe~\underline{\textbf{Project Page}}}\hspace{2em}
  \href{https://github.com/peter-lizijia/nfwm}{\faGithub~\underline{\textbf{GitHub}}}\hspace{2em}
  \href{https://huggingface.co/datasets/PeterLiZijia/nfwm}{\faDatabase~\underline{\textbf{Hugging Face}}}}}

\begin{document}

\raggedbottom
\clubpenalty=10000
\widowpenalty=10000
\displaywidowpenalty=10000

\maketitle
\thispagestyle{firstpagestyle}

\begin{abstract}
A world model lets an agent anticipate the consequences of its actions and use imagined futures to make decisions. We introduce Normalizing-Flow World Model (\method), which pairs an invertible observation encoder with an action-conditioned probabilistic predictor. Because the encoder is invertible, the predicted distribution over latent states defines a distribution over future observations. Maximizing the exact conditional likelihood of those observations therefore trains the representation and dynamics jointly. An agent can plan directly in the learned latent space and decode sampled trajectories to inspect possible futures. On Teleport, these samples capture the two possible outcomes of the same action. In Flappy Bird, a world model trained by likelihood alone supports reward-based planning that reaches 100 pipes in all ten evaluated games. For goal-conditioned control, an auxiliary inverse-dynamics objective shapes the coordinates used to compare predicted states with a goal. On LIBERO, \method uses this distance to select successful action sequences more often than LeWM and SMWM. In Atari Boxing, training on imagined trajectories improves a policy beyond its behavior-cloned initialization. Together, these results establish normalizing flows as capable world models.
\end{abstract}

\noindent\begin{minipage}{\linewidth}
  \centering
  \setlength{\parskip}{0pt}
  \captionsetup{type=figure,font=footnotesize,skip=4pt}
  \includegraphics[width=\linewidth]{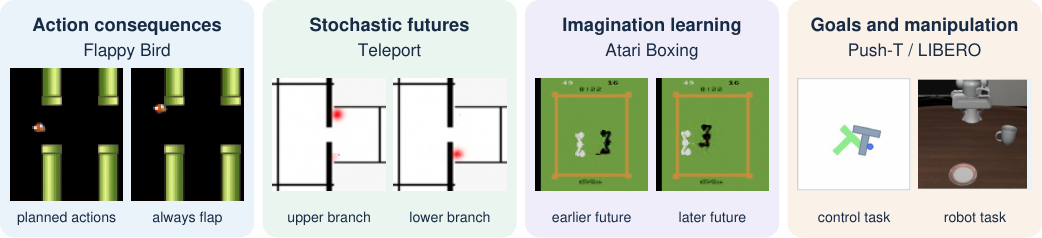}\par
  \caption{\textbf{Normalizing flows as capable world models.} Complementary case studies test action consequences, stochastic futures, imagination learning, and latent-distance decisions. Flappy Bird, Teleport, and Boxing panels show decoded predictions. Push-T and LIBERO images illustrate the control settings. Tasks use separately trained models and task-specific decision procedures.}
  \label{fig:nfwm-applications}
\end{minipage}

\begingroup
\titlespacing*{\section}{0pt}{0pt}{0.3\baselineskip}
\input{main/1intro.tex}
\endgroup
\input{main/3background.tex}
\input{main/4method.tex}
\input{main/5using.tex}
\Needspace{6\baselineskip}
\suppressfloats[t]
\input{main/6experiments.tex}
\input{main/7discussion.tex}
\input{main/2related_works.tex}
\Needspace{6\baselineskip}
\input{main/7conclusion.tex}

\newpage
\input{main/9appendix.tex}

\end{document}

%% file: main/1intro.tex
\section{Introduction}
\label{sec:introduction}

A world model lets an agent ask what would happen under different actions before executing them. Its predictions can be used to compare candidate trajectories or to train a policy from imagined experience~\citep{ha2018world,hafner2018planet,hafner2019dream}. For these uses, a model must predict action-dependent changes, carry those predictions forward over time, and expose information that a planner or policy can use. When outcomes are uncertain, it should represent different possible futures. Decoding observations makes these predictions available for inspection, while latent prediction allows decisions to be made without rendering every imagined frame.

These uses raise three questions about the state in which a world model makes predictions. \emph{Can the representation and stochastic dynamics be learned jointly, so that the model learns the coordinates in which it predicts?} \emph{Can a state that preserves the information in an observation also support decisions?} \emph{Can a planner or policy use the predicted states without decoding every future image?}

One way to connect latent prediction with observation generation is to make the observation encoder invertible. A distribution over latent states then defines a distribution over observations, and a sampled state can be mapped back to an image. Normalizing flows provide such a mapping and allow the observation likelihood to be evaluated exactly~\citep{dinh2016realnvp,zhai2025normalizing}. To turn this construction into a world model, we need to learn how the distribution of the next state depends on the observed history and the agent's action, and how to use those predictions to choose among actions.

We introduce Normalizing-Flow World Model (\textbf{\method}), which jointly learns an invertible observation encoder and an action-conditioned probabilistic predictor through exact conditional likelihood. The encoder's change of variables turns the density of future latent states into an observation density, allowing the representation and dynamics to be learned together without a separate anti-collapse regularizer. A planner can score the predicted states with a task readout or compare them with an encoded goal. The model rolls forward in latent space and maps sampled futures back to observations through the inverse encoder. By combining observation generation with direct latent-space prediction and decision making, \method bridges the capabilities of video models and latent world models.

Generative latent world models learn stochastic dynamics for planning and policy learning in imagination~\citep{hafner2018planet,hafner2019dream,hafner2023dreamerv3}. Joint-embedding predictive architectures (JEPAs) learn predictive representations for control~\citep{assran2025vjepa2,maes2026leworldmodel,ivashkov2026smwm}, including diffusion-based latent prediction in JEDI~\citep{lim2026jedi}. We compare \method with LeWM, SMWM, JEDI, DreamerV3, and Wan, an action-conditioned video model.

We test these capabilities in several settings (Figure~\ref{fig:nfwm-applications}). In Flappy Bird, likelihood-only world-model training supports reward-based planning that reaches 100 pipes in all ten evaluated games. In goal-conditioned visual control, the planner compares predicted states with an encoded goal. Teleport tests the same approach when one action can lead to two destinations. On LIBERO, we evaluate both action selection and policies trained on frozen features. In Atari Boxing, policies trained in imagination improve over their behavior-cloned initialization in all three runs.

\begin{samepage}
Our contributions are as follows:
\begin{itemize}[leftmargin=*,topsep=3pt,itemsep=3pt,parsep=0pt,partopsep=0pt]
    \item We formulate \method, jointly learning an invertible representation and action-conditioned probabilistic dynamics through exact maximum likelihood.
    \item We show how its latent predictions support observation generation, reward-based planning, policy learning in imagination, and goal-distance planning, with task-specific geometry shaping where needed.
    \item We evaluate these capabilities through complementary visual-control and manipulation case studies, and analyze conditional expressivity, predictive-state size, rollout scoring, decision geometry, and computational cost.
\end{itemize}
\end{samepage}

%% file: main/3background.tex
\section{Background}
\label{sec:background}

\paragraph{Normalizing flows}
A normalizing flow models observations through an invertible transformation $f_\theta$ with a tractable Jacobian determinant. For a base density $p_0$, the change-of-variables formula gives
\[
    p_\theta(o)=p_0(f_\theta(o))\left|\det J_{f_\theta}(o)\right|.
\]
This supports exact maximum likelihood training and generation by drawing $z\sim p_0$ and applying $o=f_\theta^{-1}(z)$~\citep{dinh2016realnvp}. TARFlow implements the transformation with autoregressive Transformer blocks over image patches~\citep{zhai2025normalizing}. Following TARFlow, we add Gaussian observation noise for density learning. Exact-likelihood statements refer to this noise-smoothed observation distribution. The invertible latent state has the same dimensionality as the observation.

\paragraph{Action-conditioned world models}
We consider offline trajectories of observations $o_t$ and actions $a_t$. A world model predicts future observations or latent states given an observed history and a candidate action sequence. Write $z_t=\mathrm{Enc}_\theta(o_t)$, $e_{a_t}=\mathrm{MLP}(a_t)$, and $h_t=(z_0,a_0,\ldots,z_{t-1},a_{t-1})$ for the encoded observation, action embedding, and preceding history. A probabilistic latent predictor models
\[
    p_\phi(z_{t+1}\mid z_t,e_{a_t},h_t).
\]
Sampled predictions enter the history recursively during imagined rollout. A planner evaluates candidate trajectories, while a policy can be trained from imagined experience~\citep{hafner2018planet,hafner2019dream}.

\paragraph{Predictive representations}
Generative latent world models learn representations together with predictive distributions~\citep{ha2018world,hafner2023dreamerv3}. JEPAs learn representations through prediction in latent space~\citep{assran2023ijepa,assran2025vjepa2}. For example, the LeWM and SMWM configurations used in our comparisons predict a point $\hat z_{t+1}$ and minimize its mean squared error (MSE) relative to the encoded next observation. Joint regression admits a constant representation, so LeWM uses SIGReg and SMWM uses inverse dynamics to constrain representation learning~\citep{maes2026leworldmodel,ivashkov2026smwm}. A JEPA can also use a distributional predictor, as in JEDI~\citep{lim2026jedi}. The representation-learning framework and the choice of predictive distribution are separate decisions.

%% file: main/4method.tex
\section{Normalizing Flows as World Models}
\label{sec:method}
\label{sec:problem-formulation}
\label{sec:nfwm-objective}

\begin{figure}[t]
    \centering
    \begin{minipage}{0.80\linewidth}
        \centering
        \makebox[0.5\linewidth]{\sffamily\footnotesize (a) LeWM~\citep{maes2026leworldmodel}}%
        \makebox[0.5\linewidth]{\sffamily\footnotesize (b) \method (Ours)}\par
        \vspace{2pt}
        \includegraphics[width=\linewidth]{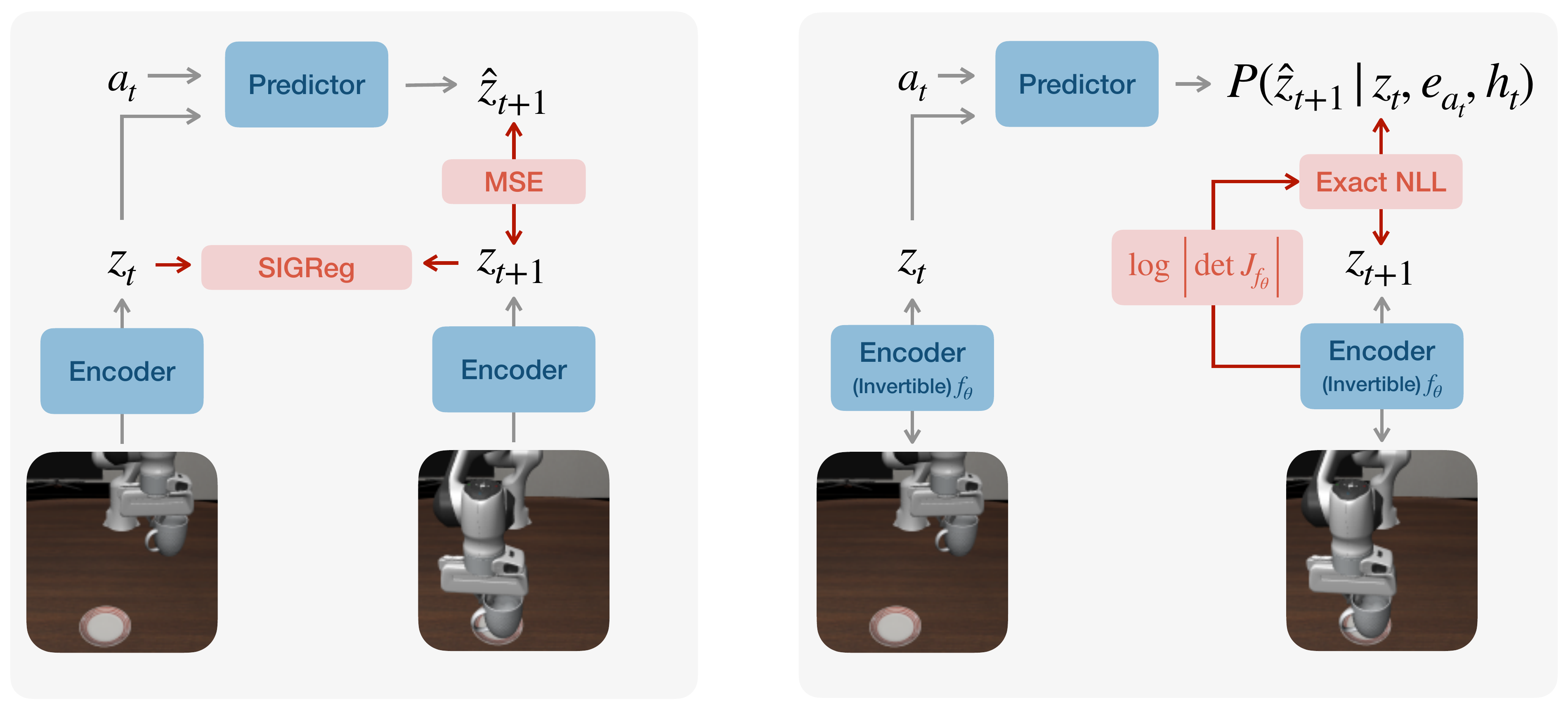}
    \end{minipage}
    \caption{\textbf{Learning a predictive representation.} (a) LeWM uses latent regression and SIGReg. (b) \method jointly learns an invertible encoder and a conditional predictive density through exact observation likelihood.}
    \label{fig:nfwm-pipeline}
\end{figure}

An observation flow becomes a world model by replacing its fixed base density with a predictive density conditioned on history and actions (Figure~\ref{fig:nfwm-pipeline}). We first describe this construction, then the predictor used in our main benchmark configurations.

\subsection{An invertible state space and a conditional density}
\label{sec:nfwm-model}

The encoder $\mathrm{Enc}_\theta$ maps an observation to latent patch tokens. For notation, we flatten them into a vector $z_t\in\mathbb R^d$:
\[
    z_t=\mathrm{Enc}_\theta(o_t),\qquad
    o_t=\mathrm{Enc}_\theta^{-1}(z_t).
\]
The predictor models $p_\phi(z_{t+1}\mid z_t,e_{a_t},h_t)$. Because the observation mapping is invertible, this defines a conditional observation density:
\begin{equation}
    p_{\theta,\phi}(o_{t+1}\mid o_{\leq t},a_{\leq t})
    =p_\phi(z_{t+1}\mid z_t,e_{a_t},h_t)
    \left|\det J_{\mathrm{Enc}_\theta}(o_{t+1})\right|.
    \label{eq:conditional-observation-density}
\end{equation}
The history is encoded with the same encoder. Its coordinates are learned together with the predictor, while invertibility retains the observation information.

\subsection{Joint exact-likelihood training}

We minimize the conditional negative log-likelihood (NLL) of the next observation:
\begin{equation}
    \mathcal L_{\mathrm{nll}}
    =-\log p_\phi(z_{t+1}\mid z_t,e_{a_t},h_t)
    -\log\left|\det J_{\mathrm{Enc}_\theta}(o_{t+1})\right|.
    \label{eq:generative-objective}
\end{equation}
Gradients update both the encoder and predictor. The first term fits the predictive density in the encoder's coordinates. The Jacobian term accounts for the change in volume under the encoder, and invertibility rules out a constant representation.

\subsection{Predictive distributions and temporal rollout}
\label{sec:training-inference}

Our primary predictor is a Transformer that outputs a diagonal Gaussian:
\begin{equation*}
    \begin{aligned}
        (\mu_{t+1},\sigma_{t+1})&=\mathrm{Pred}_\phi(z_t,e_{a_t},h_t),\\
        p_\phi(z_{t+1}\mid z_t,e_{a_t},h_t)
        &=\mathcal N\!\left(z_{t+1}\mid\mu_{t+1},
            \operatorname{diag}(\sigma_{t+1}^{2})\right).
    \end{aligned}
\end{equation*}
Each component of $\sigma_{t+1}$ is positive. The latent NLL is
\[
    -\log p_\phi
    =\frac12\sum_j\frac{(z_{t+1,j}-\mu_{t+1,j})^2}{\sigma_{t+1,j}^{2}}
    +\sum_j\log\sigma_{t+1,j}+\mathrm{const}.
\]
We sample the next state as
\begin{equation}
    \hat z_{t+1}=\mu_{t+1}+\sigma_{t+1}\odot\epsilon_t,
    \qquad \epsilon_t\sim\mathcal N(\mathbf 0,I_d),
    \label{eq:latent-sampling}
\end{equation}
and feed it into the history window for the next prediction. This yields an action-conditioned trajectory without decoding intermediate states.

The primary configuration predicts all next-frame patch parameters in parallel. A token-wise variant instead conditions each predicted patch on previously generated patches in that frame. Both are autoregressive over time, but the latter also models dependencies within a frame. Section~\ref{sec:probabilistic-dynamics} compares the two predictors.

\subsection{Inspecting imagined observations}
\label{sec:generation}

The inverse encoder maps a sampled state back to an observation:
\[
    \hat o_{t+1}=\mathrm{Enc}_\theta^{-1}(\hat z_{t+1}).
\]
We add Gaussian noise to observations during training, so decoded samples also contain observation noise. To reduce this noise in visualizations, we apply a conditional score-denoising step following TARFlow~\citep{zhai2025normalizing}, with no additional training (Appendix~\ref{app:rollout-visualization}). Decoding and score denoising are optional visualization steps. Planning and imagination learning operate directly on latent states and model features.

%% file: main/5using.tex
\section{Predicting and Acting with \method}
\label{sec:using-nfwm}

The predictive density supplies imagined trajectories. We score candidate action sequences in two ways. Task readouts estimate the return along a trajectory, while a latent goal distance measures how closely its predicted endpoint matches a goal observation. We also use predicted returns to train a policy in imagination.

\subsection{Planning through task readouts}

A reward or termination readout can evaluate imagined trajectories using latent states or predictor features. For a trajectory return $R$ and candidate action sequence $a_{\mathrm{traj}}$ of horizon $H$, the reward-based planning objective is
\begin{equation}
    a_{\mathrm{traj}}^*=\arg\max_{a_{\mathrm{traj}}}
    \mathbb E_{p_{\theta,\phi}}\!\left[
        R(\hat z_{t+1:t+H},a_{\mathrm{traj}})
    \right].
    \label{eq:reward-planning}
\end{equation}
Sampled rollouts provide scores for an optimizer such as the cross-entropy method (CEM). This formulation uses task readouts rather than assuming that Euclidean latent distance measures task progress.

In Flappy Bird, we first train the world model by likelihood alone. We then freeze its encoder and predictor and fit task readouts on predictor hidden states. CEM ranks action sequences by a survival-weighted score combining a per-step survival bonus with predicted reward (Appendix~\ref{app:flappy-planning}). The agent executes the first action and replans from the next observation.

\subsection{Planning with latent goal distances}
\label{sec:latent-geometry}
\label{sec:inverse-dynamics}

\paragraph{Goal-distance scoring}
Following LeWM~\citep{maes2026leworldmodel}, we encode the goal observation $o_g$ as $z_g=\mathrm{Enc}_\theta(o_g)$. For each candidate, we sample intermediate states, then score the final conditional mean:
\begin{equation}
    a_{\mathrm{traj}}^*=\arg\min_{a_{\mathrm{traj}}}
    \left\|\mu_{\mathrm{final}}(a_{\mathrm{traj}})-z_g\right\|_2^2.
    \label{eq:planning-objective}
\end{equation}
This is the \emph{Final Mean} protocol used in the main goal-conditioned and LIBERO experiments. It uses one rollout per candidate and omits the terminal variance term in expected squared distance. Appendix~\ref{app:rollout-scoring} gives the exact protocol, and Appendix~\ref{app:rollout-budget-followup} compares rollout counts and scoring rules. The score can guide CEM or rank a fixed candidate pool.

\paragraph{Adapting the geometry}
Invertibility preserves information, but it does not determine which differences dominate Euclidean distance. For goal-distance planning, we add inverse-dynamics supervision to encourage sensitivity to action-induced changes~\citep{pan2022isodream,ivashkov2026smwm,zhang2026delta}. For continuous actions, an inverse-dynamics model (IDM) predicts the action embedding from consecutive observed latent states:
\[
    \mathcal L_{\mathrm{inv}}=
    \left\|\mathrm{IDM}_\psi(z_t,z_{t+1})
        -\operatorname{sg}(e_{a_t})\right\|_2^2.
\]
The target action embedding is detached. A learned query readout and MLP implement this model (Appendix~\ref{app:model-training}). Where this auxiliary objective is used, training minimizes
\[
    \mathcal L=\mathcal L_{\mathrm{nll}}+
    \lambda_{\mathrm{inv}}\mathcal L_{\mathrm{inv}}.
\]
Setting $\lambda_{\mathrm{inv}}=0$ recovers likelihood-only world-model training. Section~\ref{sec:ablations} examines how inverse dynamics shapes the learned coordinates.

\subsection{Learning policies in imagination}

A policy can learn from imagined trajectories instead of optimizing a fresh action sequence at each decision. Following Dreamer~\citep{hafner2019dream,hafner2025training}, we train an actor--critic on model rollouts with predicted rewards and termination. In Atari Boxing, task heads are fitted to frozen world-model features and the actor starts from behavior cloning on the offline data. Imagination training updates the policy using predicted experience. During evaluation, the policy receives a new observation at each decision (Section~\ref{sec:stochastic-decisions}).

%% file: main/6experiments.tex
\section{Experiments}
\label{sec:experiments}

We evaluate \method on reward-based planning in Flappy Bird, goal-conditioned visual control, and stochastic planning in Teleport. We then evaluate action selection and frozen-encoder action experts on LIBERO, and train policies on imagined trajectories in Atari Boxing. We train world models separately across benchmarks.

\paragraph{Comparators and scope}
LeWM~\citep{maes2026leworldmodel} and SMWM~\citep{ivashkov2026smwm} use approximately 5M-parameter encoders and 10M-parameter predictors, with compact frame embeddings. We also evaluate full-patch versions that match \method's predictive-state size (Appendix~\ref{app:patch-state-followup}). Boxing includes offline JEDI~\citep{lim2026jedi} and DreamerV3~\citep{hafner2023dreamerv3}, with DreamerV3's memory restricted to three frames. We fine-tune a 1.3B-parameter autoregressive Wan model for action-conditioned video prediction and evaluate it at 1, 2, and 4 function evaluations (NFE) per latent. We score its plans by goal distance in the latent space of its variational autoencoder (VAE) (Appendix~\ref{app:wan}).

\subsection{Reward-based closed-loop planning}
\label{sec:reward-based-planning}

\paragraph{Environment}
We use Flappy Bird to test reward-based planning from visual observations. The agent chooses between flap and no-op to pass through pipes while avoiding collisions.

\begin{figure}[t]
    \centering
    \includegraphics[width=0.82\linewidth]{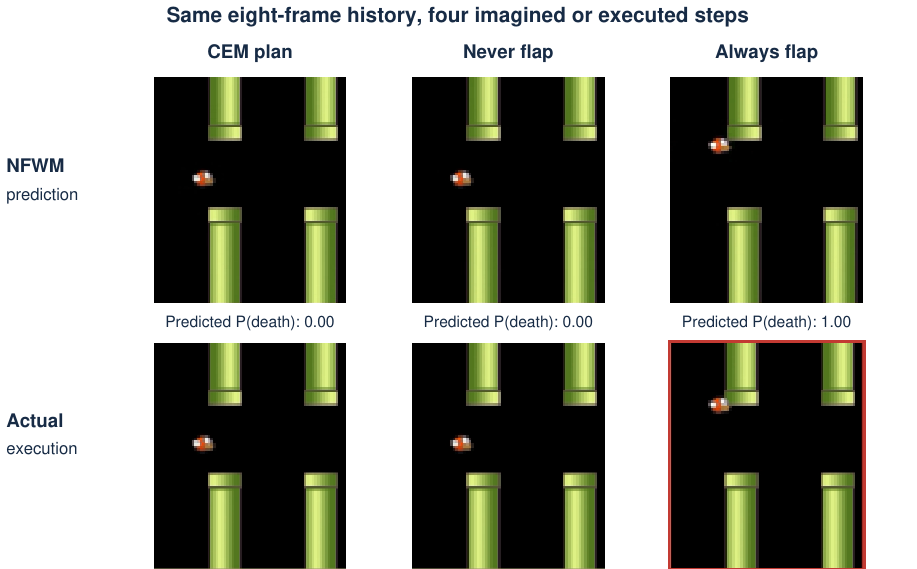}
    \caption{\textbf{Action-conditioned predictions on Flappy Bird.} Sampled predictions (top) and environment outcomes (bottom) for three action sequences from the same history. The red border marks collision under always-flap actions.}
    \label{fig:flappy-action-consequences}
\end{figure}

\paragraph{Predicting action consequences}
Figure~\ref{fig:flappy-action-consequences} compares a planned action sequence with never-flap and always-flap actions from the same observation history. The top row shows model predictions, and the bottom row shows the corresponding environment outcomes. The predictions distinguish the surviving plan from collision-inducing actions, supplying a signal for ranking trajectories before execution.

\paragraph{Model and planning}
We train the world model by exact observation likelihood with $\lambda_{\mathrm{inv}}=0$, then freeze it and fit reward and termination readouts to the predictor's hidden states. CEM ranks trajectories by a survival-weighted reward score and replans after each executed action. Appendix~\ref{app:flappy-bird} gives the data, architecture, training settings, and planning protocol.

\begin{table}[htbp]
    \captionsetup{position=top,skip=6pt}
    \centering\footnotesize
    \caption{\textbf{Closed-loop planning on Flappy Bird.} Results on ten held-out levels, with a stopping threshold of 100 pipes. All models use readouts fitted to frozen predictor hidden states and the same CEM budget.}
    \label{tab:flappy-planning}
    \begin{tabular*}{0.90\linewidth}{@{\extracolsep{\fill}}lrr@{}}
        \toprule
        Model & Games reaching 100 pipes & Best pipes \\
        \midrule
        \method, likelihood only & $10/10$ & 100 \\
        JEPA-style, invertible encoder & $0/10$ & 1 \\
        JEPA-style, ViT encoder & $0/10$ & 0 \\
        \bottomrule
    \end{tabular*}
\end{table}

\paragraph{Results}
On ten level seeds excluded from training, the model reaches the 100-pipe stopping threshold in every game. Table~\ref{tab:flappy-planning} compares this result with JEPA-style controls trained with latent MSE and SIGReg. The invertible-encoder control clears at most one pipe, and the ViT-encoder control clears none. The result shows that likelihood-trained predictive features support closed-loop control through readouts fitted afterwards.

\subsection{Goal-conditioned visual control}
\label{sec:goal-conditioned-planning}

We next test latent goal-distance planning on Push-T, Reacher, Cube, and TwoRoom.

\begin{figure}[htbp]
    \centering
    \includegraphics[width=\linewidth]{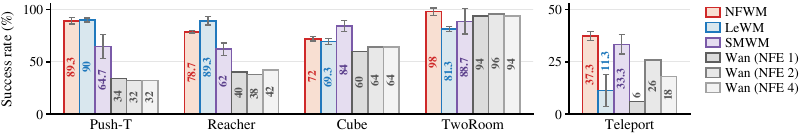}
    \caption{\textbf{Planning performance across environments.} Mean $\pm$ std over three training runs for \method, LeWM, and SMWM. Wan uses one training run per task. Averaged over the four deterministic tasks, \method, LeWM, and SMWM achieve 84.5\%, 82.5\%, and 74.8\% success, respectively.}
    \label{fig:planning-success}
\end{figure}

\paragraph{Environment and data}
We follow the four deterministic visual-control tasks and offline data setup of LeWM~\citep{maes2026leworldmodel}. Appendix~\ref{app:datasets} describes the tasks and data.

\paragraph{Planning and results}
For \method, CEM~\citep{rubinstein1999crossentropy} optimizes Equation~\ref{eq:planning-objective} with one rollout per candidate and Final Mean scoring. Each plan spans 25 environment steps, with replanning allowed within a total budget of 50 steps. Figure~\ref{fig:planning-success} reports success on these tasks and on Teleport (Section~\ref{sec:teleport-case}). \method performs comparably to LeWM on Push-T, between LeWM and SMWM on Reacher and Cube, and best among these configurations on TwoRoom. It has the highest four-task average in this comparison, 84.5\%. With inverse dynamics shaping the coordinates, its probabilistic predictions also provide a useful goal-distance signal in deterministic tasks.

\subsection{Stochastic prediction and open-loop planning}
\label{sec:teleport-case}

\begin{figure}[htbp]
    \centering
    \begin{minipage}[c]{0.28\linewidth}
        \includegraphics[width=\linewidth]{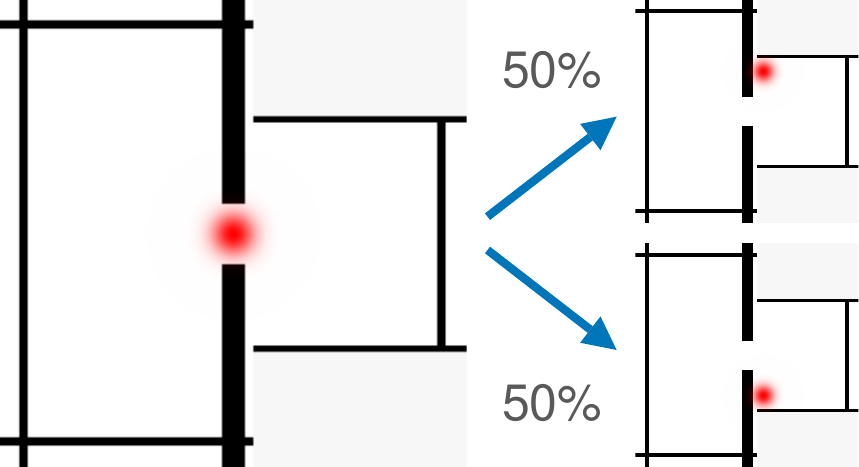}
    \end{minipage}\hspace{0.04\linewidth}%
    \begin{minipage}[c]{0.64\linewidth}
        \caption{\textbf{Teleport.} Entering the doorway selects the upper or lower destination with equal probability.}
        \label{fig:teleport-transition}
    \end{minipage}
\end{figure}

\paragraph{Task and planning}
Teleport modifies TwoRoom so that entering a doorway sends the agent to one of two destinations with equal probability (Figure~\ref{fig:teleport-transition}). We optimize the latent goal-distance score with CEM and execute a single 25-step plan without replanning. The paired cases have identical start and goal images, so a plan tailored to one destination encounters it with probability 50\% (Appendix~\ref{app:datasets}).

\begin{figure}[htbp]
    \centering
\input{figures/teleport_analysis_compact.tex}
    \caption{\textbf{Predictor rollouts on Teleport.} We visualize decoded latent states predicted from context observations and action sequences. Step $t+5$ shows the agent just after it passes through the doorway. For LeWM and SMWM, we train a decoder with the encoder frozen. For \method, we invert the encoder.}
    \label{fig:teleport-prediction-analysis}
\end{figure}
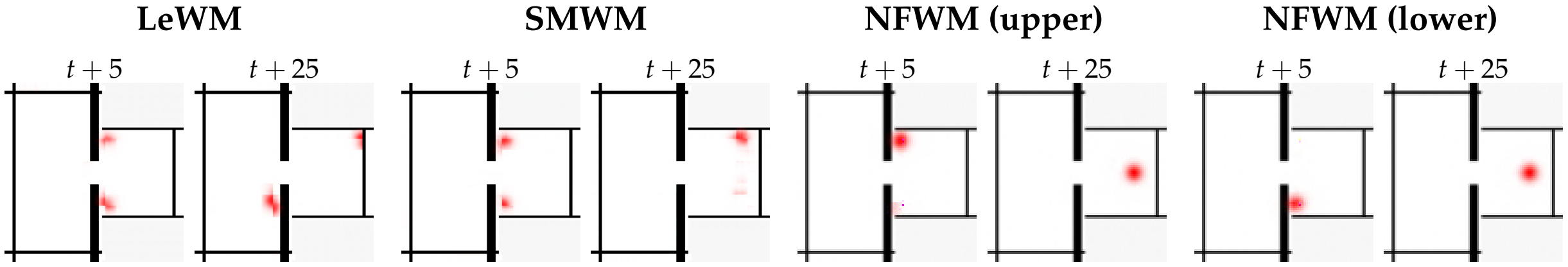

\paragraph{Predictions and task success}
Figure~\ref{fig:teleport-prediction-analysis} shows \method's distinct continuations from the same initial context and action block, preserving alternative outcomes for planning. Planning success is $37.3\pm2.3\%$, compared with $11.3\pm7.6\%$ for LeWM and $33.3\pm4.6\%$ for SMWM (Figure~\ref{fig:planning-success}).

\paragraph{Token-wise prediction}
Following TARFlow's token-wise autoregressive design~\citep{zhai2025normalizing}, we condition each next-frame patch on those already generated, giving the predictor a more expressive model of within-frame dependencies. In a single run with sampled terminal states, it reaches the goal in 25 of 50 cases (50\%) (Appendix~\ref{app:tokenwise-prediction}). Sequential generation also increases planning time (Appendix~\ref{app:compute-followup}).

\subsection{Robotic action selection and action experts}
\label{sec:action-selection}

\begin{figure}[t]
    \centering
    \includegraphics[width=0.55\linewidth]{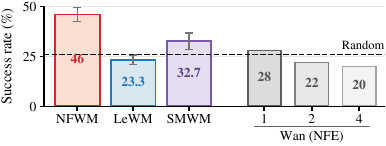}
    \caption{\textbf{Action selection performance on LIBERO.} Mean $\pm$ std over three training runs for \method, LeWM, and SMWM. Wan uses one training run. The dashed line shows uniform random selection from the same candidate pool.}
    \label{fig:libero-success}
\end{figure}

\begingroup
\widowpenalty=10000
\paragraph{Environment and data} Our dataset is collected in LIBERO-PRO~\citep{zhou2025liberopro}, using a custom version of the LIBERO task in which a robot places a white mug on a plate~\citep{liu2023libero}. The world model receives images from one external camera and the robot's actions. We train on successful and unsuccessful trajectories collected with the pretrained policy $\pi_{0.5}$~\citep{intelligence2025pi05}. Evaluation uses 50 scenes with 50 candidate action sequences per scene (Appendix~\ref{app:libero-selection}).
\par
\endgroup

\paragraph{Setup} We evaluate whether a world model can support robotic action selection on LIBERO. We generate candidate action sequences using $\pi_{0.5}$ and perturb them with noise. Each latent world model scores every candidate with Equation~\ref{eq:planning-objective} and selects the lowest-cost one. The selected sequence is executed in full without replanning.

\paragraph{Results} Figure~\ref{fig:libero-success} reports action-selection success. \method achieves $46.0\pm3.5\%$ success, compared with $23.3\pm2.3\%$ for LeWM and $32.7\pm4.2\%$ for SMWM. Wan achieves 28\%, 22\%, and 20\% with 1, 2, and 4 NFE, respectively.

With the same candidate pools, \method selects successful actions more often. Section~\ref{sec:decision-objectives} examines how predicted endpoints affect the ranking.

\paragraph{Using the representation in an action expert}
Following the frozen-encoder protocol of TC-LeWM~\citep{liu2026tclewm}, we train four action experts on \method features, one per LIBERO suite, covering 40 manipulation tasks. Each expert predicts a short action chunk and conditions the next chunk on a new observation. The experts achieve 82.7\% overall success and 70.8\% on the Long suite (Table~\ref{tab:action-expert-followup}). These results show that frozen \method representations support closed-loop policy execution across diverse tasks, including longer-horizon interactions. Appendix~\ref{app:action-expert-followup} gives the training and evaluation settings.

\begin{table}[htbp]
    \centering\footnotesize
    \caption{\textbf{LIBERO action experts with frozen \method features.} One expert per suite, evaluated on 50 trials for each of ten tasks.}
    \label{tab:action-expert-followup}
    \begin{tabular*}{0.80\linewidth}{@{\extracolsep{\fill}}lrrrrr@{}}
        \toprule
        Suite & Spatial & Object & Goal & Long & Overall \\
        \midrule
        Success (\%) & 73.8 & 96.2 & 90.0 & 70.8 & 82.7 \\
        \bottomrule
    \end{tabular*}
\end{table}

\subsection{Policy learning in imagination}
\label{sec:stochastic-decisions}

\paragraph{Environment and data} To evaluate policy learning under stochastic dynamics, we introduce random opponent actions into Atari Boxing. On each emulator frame, the opponent's action is replaced with probability 0.5 by a randomly selected different action. The opponent's actions are not provided to the world model. Similar to \citet{berdica2024robust}, the offline dataset combines trajectories from ten PPO~\citep{schulman2017ppo} checkpoints at different training stages (Appendix~\ref{app:boxing-data}).

\begin{figure}[htbp]
    \centering
    \includegraphics[width=\linewidth]{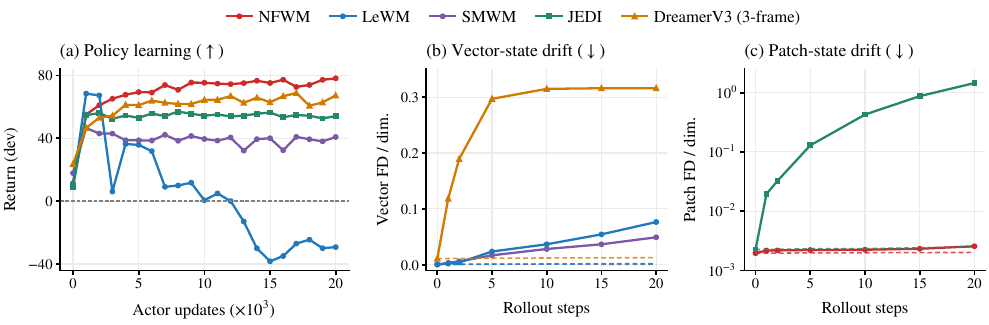}
    \caption{\textbf{Policy learning and rollout drift on Atari Boxing.} (a) Development returns averaged over three training runs, starting from the behavior-cloned policy. (b) Vector Fr\'echet distance (FD) for LeWM, SMWM, and DreamerV3. (c) Patch FD for \method and JEDI. All FD values are normalized per dimension. Solid curves show predicted futures and dashed curves show real-future controls.}
    \label{fig:boxing-policy-learning}
    \label{fig:boxing-rollout-fd}
\end{figure}

\paragraph{Setup} Following Dreamer 4~\citep{hafner2025training}, we initialize an actor through behavior cloning on the offline data, then train an actor--critic in the world model with a KL penalty to the frozen behavior-cloned policy. For \method, LeWM, and SMWM, reward and termination heads are fitted to frozen world-model features. Imagined trajectories start from encoded observations and use these heads to score future transitions. We also evaluate an offline CNN adaptation of JEDI~\citep{lim2026jedi} with three-frame conditional diffusion, and DreamerV3 with its recurrent state rebuilt from the latest three frames and actions at each decision. Appendix~\ref{app:boxing-model-training} describes both architectures and their training settings. Policies receive actual observations at every environment decision. We report mean return over 50 episodes with sampled actions (Appendix~\ref{app:atari-evaluation}).

\paragraph{Results}
\method achieves the highest mean return, $77.37\pm9.28$ (Table~\ref{tab:boxing-results}), and improves over its behavior-cloned initialization in all three runs (Table~\ref{tab:boxing-per-seed}).

\begin{table}[htbp]
    \captionsetup{position=top,skip=6pt}
    \centering\footnotesize
    \renewcommand{\arraystretch}{1.05}
    \caption{\textbf{Policy-learning returns on Atari Boxing.} Final policies evaluated on 50 held-out episodes. Mean $\pm$ std over three training runs, except Wan, which uses one run.}
    \label{tab:boxing-results}
    \begin{tabular*}{0.75\linewidth}{@{\extracolsep{\fill}}lr@{}}
        \toprule
        Model & Return \\
        \midrule
        \method & $77.37\pm9.28$ \\
        LeWM & $-35.19\pm23.97$ \\
        SMWM & $40.78\pm38.34$ \\
        JEDI & $53.92\pm6.35$ \\
        DreamerV3 (3-frame memory) & $64.47\pm6.32$ \\
        \midrule
        Wan (NFE 1) & $45.1$ \\
        Wan (NFE 2) & $59.9$ \\
        Wan (NFE 4) & $63.1$ \\
        \bottomrule
    \end{tabular*}
\end{table}

In Figure~\ref{fig:boxing-policy-learning}(a), the early gains for LeWM and SMWM diminish with further training. JEDI and DreamerV3 retain their gains, and \method continues to improve. Section~\ref{sec:boxing-analysis} relates these curves to rollout drift in Figure~\ref{fig:boxing-rollout-fd}(b,c).

%% file: figures/teleport_analysis_compact.tex
\begingroup
\newcommand{\teleportpair}[3]{%
  \begin{minipage}[t]{0.235\linewidth}
    \centering\small\textbf{#1}\par\vspace{2pt}
    \begin{minipage}[t]{0.485\linewidth}\centering
      {\scriptsize $t+5$}\par
      #2
    \end{minipage}\hfill
    \begin{minipage}[t]{0.485\linewidth}\centering
      {\scriptsize $t+25$}\par
      #3
    \end{minipage}
  \end{minipage}%
}
\teleportpair{LeWM}{\includegraphics[width=\linewidth]{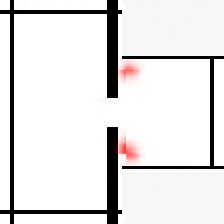}}{\includegraphics[width=\linewidth]{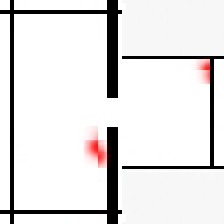}}\hfill
\teleportpair{SMWM}{\includegraphics[width=\linewidth]{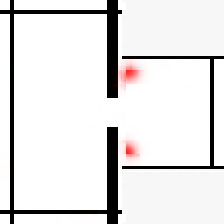}}{\includegraphics[width=\linewidth]{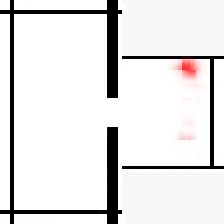}}\hfill
\teleportpair{NFWM (upper)}{\includegraphics[width=\linewidth]{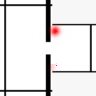}}{\includegraphics[width=\linewidth]{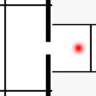}}\hfill
\teleportpair{NFWM (lower)}{\includegraphics[width=\linewidth]{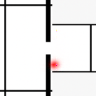}}{\includegraphics[width=\linewidth]{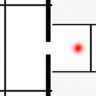}}
\endgroup

%% file: main/7discussion.tex
\section{Discussion}
\label{sec:discussion}

\subsection{What the flow construction provides}

\begin{figure}[t]
    \centering
    \includegraphics[height=0.35\textwidth,trim=0 364.5bp 204.5bp 0,clip]{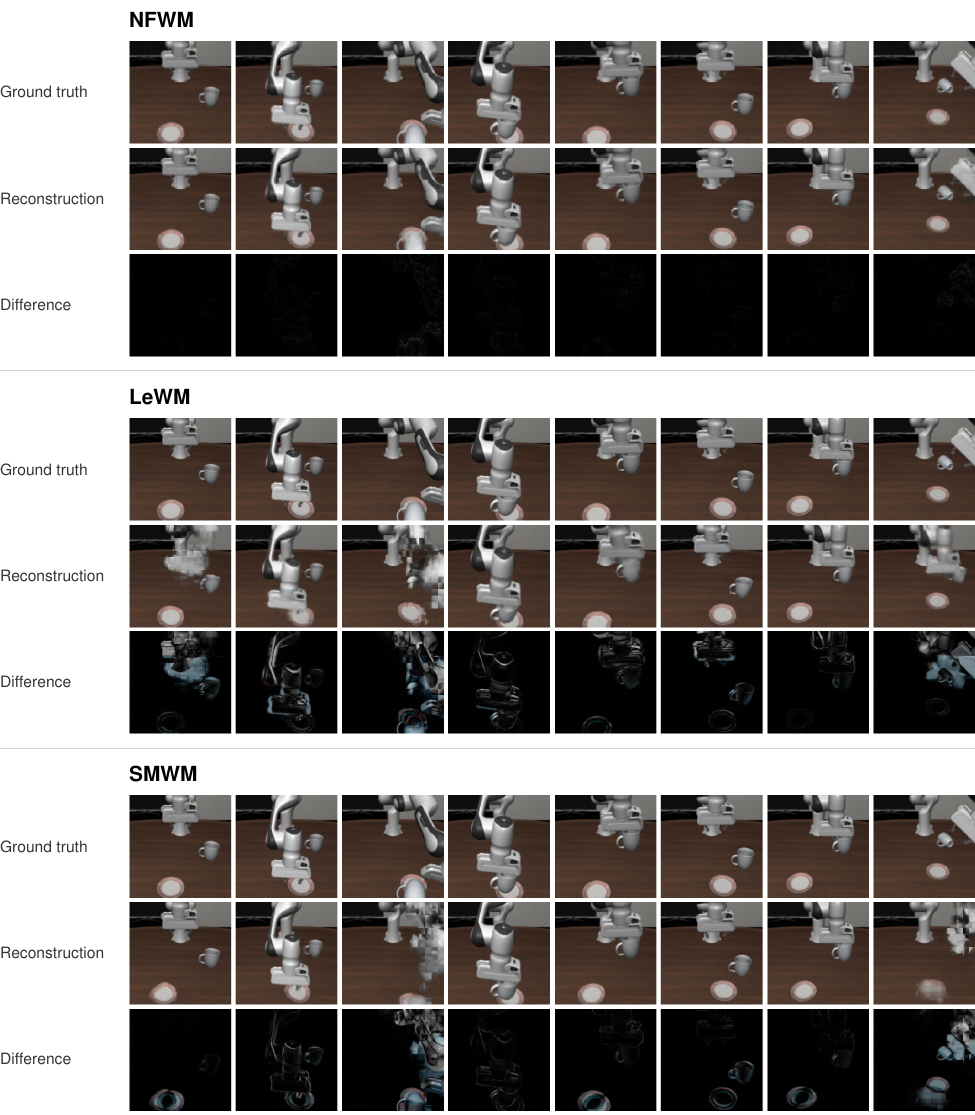}\hspace{1em}%
    \includegraphics[height=0.35\textwidth,trim=61.5bp 183.5bp 204.5bp 181bp,clip]{figures/libero_reconstructions.pdf}
    \caption{\textbf{Image reconstructions on LIBERO.} We show observations reconstructed from their encoded latent states and their differences from the original observations. Decoders are trained with the encoders frozen.}
    \label{fig:libero-recon-main}
\end{figure}

The flow construction jointly learns an information-preserving state space and a stochastic predictive density. Predicted trajectories can be used directly for goal comparison or decoded into observations. For Euclidean goal-distance control, inverse dynamics adapts the coordinates. For reward-based planning and imagination learning, reward and termination readouts score trajectories using the learned features. On LIBERO, reconstructions from \method's frozen features show finer gripper and object details than those from LeWM and SMWM (Figure~\ref{fig:libero-recon-main} and Appendix~\ref{app:libero-reconstruction}).

\subsection{Geometry for goal comparison}
\label{sec:ablations}

An invertible representation retains observation information, while its coordinates determine which differences dominate Euclidean distance. Rescaling one latent axis can preserve exact reconstruction yet change a candidate ranking. We evaluate how inverse dynamics shapes these coordinates on Push-T.

\begin{table}[htbp]
    \captionsetup{position=top,skip=6pt,width=0.60\linewidth}
    \centering\footnotesize
    \renewcommand{\arraystretch}{0.95}
    \setlength{\tabcolsep}{4pt}
    \caption{\textbf{Objective configurations.} Push-T planning success (\%).}
    \label{tab:objective-ablations}
    \begin{tabular*}{0.60\linewidth}{@{\extracolsep{\fill}}lr@{}}
        \toprule
        Variant & Success (\%) \\
        \midrule
        \method & 86.0 \\
        w/o InvDyn & 34.0 \\
        w/o NLL, clean input & 28.0 \\
        w/o NLL, noisy input & 40.0 \\
        \bottomrule
    \end{tabular*}
\end{table}

Table~\ref{tab:objective-ablations} compares one training run per configuration. Removing inverse dynamics reduces Push-T goal-distance planning success from 86\% to 34\%. Replacing NLL with forward MSE gives 28\% with clean inputs and 40\% with noisy inputs. The forward-MSE model plans more successfully with noisy inputs, although it still falls short of the full model. MSE trains a point prediction, whereas NLL learns both the mean and predictive variance. Together, these results support combining likelihood training and inverse dynamics for Euclidean goal-distance planning. Appendix~\ref{app:ablation-details} specifies the noise and rollout settings.

\begin{figure}[t]
    \centering
    \includegraphics[width=\linewidth]{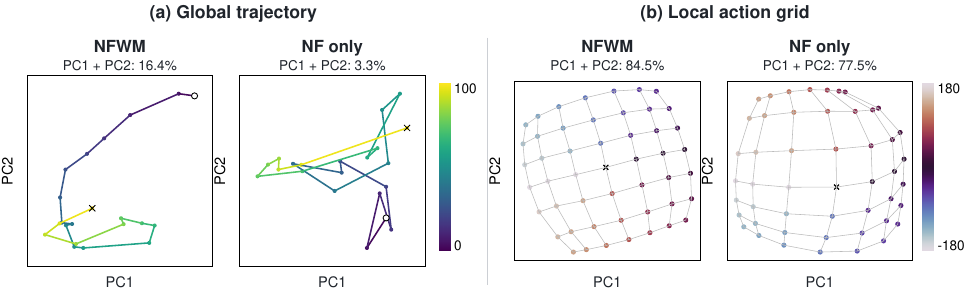}
    \caption{\textbf{Latent geometry of \method.} Principal component analysis (PCA) of latent states from \method and its variant without inverse dynamics (w/o InvDyn). (a) A held-out trajectory, colored by time. The circle and cross mark its start and end. (b) Predicted states for a grid of action offsets from one state, colored by action direction.}
    \label{fig:geometry-main}
\end{figure}

Figure~\ref{fig:geometry-main} shows the first two principal components of the latent states. \method exhibits fewer abrupt turns in the trajectory projection and a more regular grid of predicted states under changes in action than the variant without inverse dynamics. Under the same PCA evaluation protocol, Wan's trajectory has sharper reversals and tightly grouped states connected by longer segments, while both models produce ordered local action grids (Figure~\ref{fig:wan-vae-pca} in Appendix~\ref{app:representation-analysis}). The less regular global trajectory suggests that Euclidean distance in Wan's VAE space may be less suited to goal comparison. The held-out probes in Appendix~\ref{app:representation-analysis} also show that action information is more accessible with inverse dynamics.

\subsection{Conditional expressivity and stochastic prediction}
\label{sec:boxing-analysis}

\paragraph{Different possible futures}
Teleport introduces two possible destinations for the same observed state and action. \method produces distinct continuations, while the LeWM and SMWM point predictors initially place predictions between the possible outcomes (Figure~\ref{fig:teleport-prediction-analysis}). Predicting a distribution allows the model to retain alternative futures for planning.

\paragraph{Rollout drift and imagination learning}
In Boxing, unobserved opponent actions create uncertainty about future observations. Figure~\ref{fig:boxing-rollout-fd}(b) tracks latent Fr\'echet distance (FD) from training states. In the comparison between deterministic models, LeWM shows larger long-horizon drift than SMWM and lower policy return. The covariance participation ratio gives an effective latent dimension of 55.59 for LeWM and 1.95 for SMWM (Appendix~\ref{app:effective-dimension}). Concentrating variation in fewer directions may help SMWM limit rollout error accumulation.

Across Figure~\ref{fig:boxing-rollout-fd}(b,c), \method remains close to its encoded real-future control as rollout length grows. JEDI's patch FD continues to increase, whereas DreamerV3's vector FD rises rapidly over the first five steps and then approaches a plateau. Both stochastic baselines retain their gains during imagination learning and outperform LeWM and SMWM. One possible explanation is that stochastic prediction preserves alternative futures useful for policy learning even when the imagined state distribution shifts. \method combines rollouts close to its real-future control with the highest mean policy return. We interpret each FD curve relative to its own real-future control because the models use different latent representations. Appendix~\ref{app:boxing-fd} defines the metrics and reports the controls.

\subsection{Predicting outcomes and ranking them}
\label{sec:decision-objectives}

\begin{figure}[htbp]
    \centering
    \includegraphics[width=\linewidth]{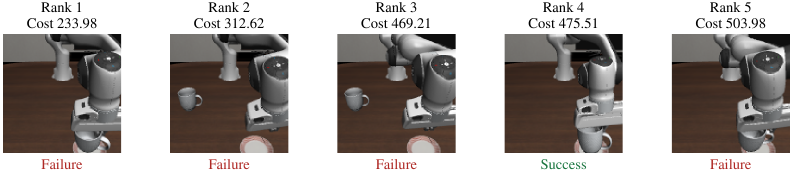}
    \caption{\textbf{Failure case on LIBERO.} Actual simulator terminal images for the five lowest-cost candidates of \method. The top-ranked candidate fails, while rank 4 succeeds.}
    \label{fig:libero-top5}
\end{figure}

LIBERO distinguishes transition prediction from goal comparison. Replacing predicted endpoints with encoded simulator endpoints improves all three models' rankings (Table~\ref{tab:libero-ranking-diagnosis}). NFWM's lead over LeWM shrinks from 22.7 to 9.3 points, and its lead over SMWM from 13.3 to 4.0 points. Thus, endpoint prediction contributes to the observed gaps. Together with the reconstructions in Figure~\ref{fig:libero-recon-main}, this suggests that \method learns a representation better suited to predicting action outcomes. Even simulator-endpoint scoring reaches only 54\% for NFWM. In Figure~\ref{fig:libero-top5}, the fourth-ranked sequence succeeds while the selected sequence fails. A candidate may succeed during execution but finish far from the goal image, so terminal distance can disagree with the task criterion even with a correctly predicted endpoint.

\begin{table}[htbp]
    \captionsetup{position=top,skip=6pt}
    \centering\footnotesize
    \renewcommand{\arraystretch}{0.95}
    \setlength{\tabcolsep}{2pt}
    \caption{\textbf{Ranking diagnosis on LIBERO (\%).} Mean $\pm$ std over three training runs. Predictor columns rank candidates by predicted terminal latent distance to the goal. Simulator Top-1 uses encoded simulator endpoints. Top-5 measures whether any of the five highest-ranked candidates succeeds.}
    \label{tab:libero-ranking-diagnosis}
    \begin{tabular*}{0.85\linewidth}{@{\extracolsep{\fill}}lccc@{}}
        \toprule
        Model & Predictor Top-1 & Predictor Top-5 & Simulator Top-1 \\
        \midrule
        \method & $46.0{\pm}3.5$ & $78.7{\pm}1.2$ & $54.0{\pm}2.0$ \\
        LeWM & $23.3{\pm}2.3$ & $73.3{\pm}1.2$ & $44.7{\pm}4.2$ \\
        SMWM & $32.7{\pm}4.2$ & $71.3{\pm}4.2$ & $50.0{\pm}4.0$ \\
        \bottomrule
    \end{tabular*}
\end{table}

\begin{samepage}
NFWM's top-five coverage is 78.7\%, compared with 46.0\% top-one success. Uniform random ranking gives 70.3\% top-five coverage for these pools (Appendix~\ref{app:libero-diagnosis}). Successful candidates often appear among the top five but are not selected first, motivating task-aware outcome scores for manipulation.
\par
\end{samepage}

\subsection{Predictive-state size and computational cost}
\label{sec:compute}
\label{sec:probabilistic-dynamics}

On an RTX PRO 6000 GPU, Push-T planning takes 10.038 s per decision for \method and 10.478 s for full-patch LeWM, compared with 0.286 s for compact LeWM. Retaining the full patch state therefore carries a computational cost. Appendix~\ref{app:compute-followup} reports prediction latency, rollout throughput, peak memory, and decision time, including the additional cost of token-wise prediction.

\paragraph{Predictive-state controls}
Matching \method's $144\times192$ predictive-state size does not consistently improve LeWM or SMWM (Appendix~\ref{app:patch-state-followup}). State size alone therefore does not explain the planning differences.

\paragraph{Parallel and token-wise prediction}
The parallel predictor generates a full latent frame at once and already supports successful planning and policy learning across the evaluated tasks. Token-wise prediction conditions each patch on preceding patches within the same frame, allowing richer within-frame dependencies (Appendix~\ref{app:tokenwise-prediction}). Under the same search budget, a Teleport planning decision takes 22.3 times as long with the token-wise predictor as with the parallel predictor (Appendix~\ref{app:compute-followup}). We therefore use parallel prediction in the main experiments to combine strong decision performance with efficient rollout.

%% file: main/2related_works.tex
\section{Related Work}
\label{sec:related-work}

\paragraph{Generative latent world models}
Generative world models use predictive latent distributions for decisions~\citep{ha2018world}. Recurrent state-space models (RSSMs) learn representations and stochastic dynamics with variational objectives, supporting planning~\citep{hafner2018planet} and policy learning in imagination~\citep{hafner2019dream,hafner2020mastering,hafner2023dreamerv3}. Dreamer 4 uses a separately trained tokenizer~\citep{hafner2025training}. NFWM learns an invertible representation and its dynamics jointly through exact conditional observation likelihood.

\paragraph{Joint-embedding predictive models}
JEPAs learn through latent prediction without observation reconstruction~\citep{assran2023ijepa,assran2025vjepa2}. LeWM, SMWM, and Delta-JEPA use latent goal distances for control~\citep{maes2026leworldmodel,ivashkov2026smwm,zhang2026delta}. JEDI extends joint-embedding prediction to stochastic latent distributions through diffusion~\citep{lim2026jedi}. Our comparisons include compact point predictors, full-patch controls, and stochastic latent models.

\paragraph{Inverse dynamics}
Iso-Dream uses inverse dynamics to separate controllable and noncontrollable dynamics~\citep{pan2022isodream}. SMWM and Delta-JEPA also use action recovery for representation learning~\citep{ivashkov2026smwm,zhang2026delta}. We use inverse dynamics to adapt goal-distance geometry and evaluate likelihood-only world-model training in reward-based planning.

\paragraph{Normalizing flows and video prediction}
Flows support exact likelihood and generation through invertible transformations~\citep{dinh2016realnvp}. VideoFlow models conditional stochastic video~\citep{kumar2019videoflow}, while Transformer-based flows cover images~\citep{zhai2025normalizing,gu2026starflow}, video~\citep{gu2025starflow}, multimodal generation~\citep{shen2026starflow2}, and latent reasoning~\citep{tu2026latentreasoningnormalizingflows}. We study action-conditioned flow dynamics whose latent trajectories support planning and policy learning, with inverse decoding for inspecting future observations.

%% file: main/7conclusion.tex
\section{Conclusion}
\label{sec:conclusion}

NFWM turns an observation flow into an action-conditioned world model by learning a conditional latent density jointly with its invertible encoder. Its predictions support planning through task readouts and goal distances, as well as policy improvement in imagination. The experiments show that likelihood-trained flow representations can support decisions directly in latent space, with inverse dynamics adapting the coordinates for Euclidean goal comparison.

\paragraph{Limitations and future work}
The evaluation uses simulated tasks and separately trained models. Training requires observation sequences paired with action labels. Retaining full patch states increases rollout cost, and more expressive token-wise predictors add sequential computation at inference. Goal-distance planning also depends on the learned geometry, motivating task-aware scores for manipulation. Extending the approach to real-world observations, larger scenes, and shared models across tasks remains future work.

%% file: main/9appendix.tex
\appendix

\section{Goal-Conditioned Planning}
\label{app:planning}

\subsection{Environments and Datasets}
\label{app:datasets}
\label{app:planning-data}

\subsubsection{Deterministic Control}
We evaluate navigation, articulated control, planar manipulation, and
three-dimensional manipulation using TwoRoom, Reacher, Push-T, and Cube.
We follow
LeWM's datasets and planning setup~\citep{maes2026leworldmodel}, with SMWM's
held-out splitting protocol~\citep{ivashkov2026smwm}. We randomly split
whole episodes into 90\% training and 10\% validation data
(Table~\ref{tab:data-splits}) and sample eligible validation windows
uniformly without replacement. Table~\ref{tab:planning-settings} lists
the window counts and goal offsets.

\begin{table}[htbp]
    \setlength{\belowcaptionskip}{6pt}
    \centering\small
    \caption{Episode-held-out partitions for deterministic
    control. Action dimension is per raw environment step.}
    \label{tab:data-splits}
    \begin{tabular}{lrrr}
        \toprule
        Task & Training episodes & Validation episodes & Action dimension \\
        \midrule
        Push-T & 16,816 & 1,869 & 2 \\
        Cube & 9,000 & 1,000 & 5 \\
        Reacher & 9,000 & 1,000 & 2 \\
        TwoRoom & 9,000 & 1,000 & 2 \\
        \bottomrule
    \end{tabular}
\end{table}

\subsubsection{Teleport}
\begin{figure}[htbp]
    \centering
    \includegraphics[width=0.72\linewidth,
        trim=160bp 248bp 176bp 128bp,clip]{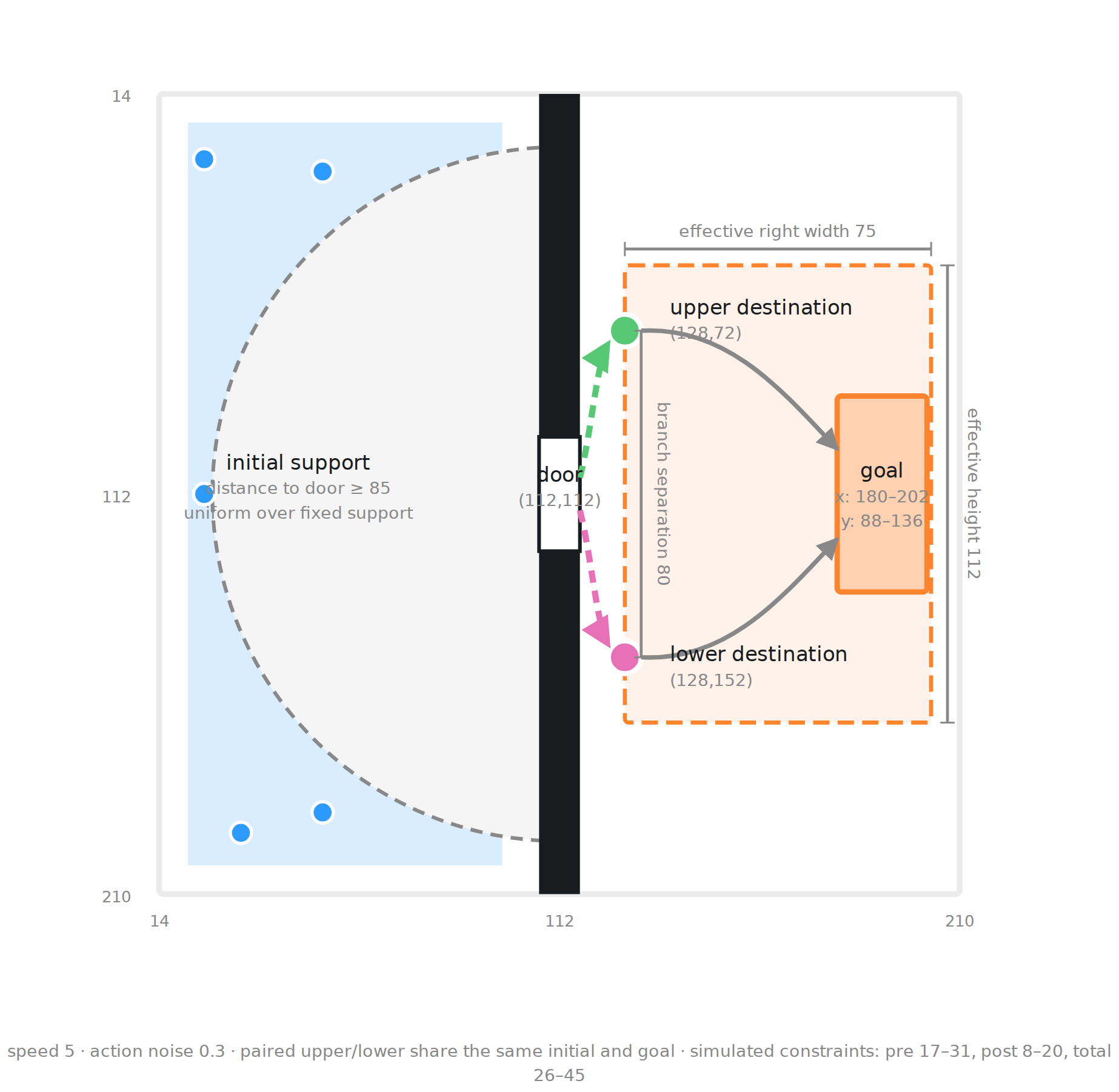}
    \caption{\textbf{Teleport geometry.} Schematic of the constrained
    initial region, doorway, and two teleport destinations. The contracted
    right room allows both branches to reach the goal within the evaluation
    horizon. The shaded goal box is the training target region, with
    $x\in[180,202]$ and $y\in[88,136]$. Paired evaluation fixes the target
    height at $y=112$.
    }
    \label{fig:teleport-construction}
\end{figure}

Teleport modifies TwoRoom so that reaching the doorway sends the agent to
one of two destinations with equal probability. We contract the right-hand
room to keep the stochastic transition within the evaluation horizon
(Figure~\ref{fig:teleport-construction}).
On the $224\times224$ canvas, the door is centered at $(112,112)$ and the
destinations are $(128,72)$ and $(128,152)$. Beyond its entrance, the right
room spans $y\in[56,168]$. Initial positions lie in $x\in[21,98]$,
$y\in[21,202]$, at least 85 pixels from the door center. Training targets
lie in $x\in[180,202]$, $y\in[88,136]$. The speed parameter is 5 and the
expert action-noise parameter is 0.5.

The training set contains 20,000 episodes: 9,000 successful trajectories arranged
as 4,500 branch pairs, 9,000 opposite-branch action replays, and 2,000
collision or exploration trajectories. Successful trajectories reach the
exact target through executed corrective actions. Opposite-branch replays
execute a recorded continuation from the other destination and retain
both successful and unsuccessful outcomes. Pairing continuations with both
branches reduces the association between expert corrections within an
action block and the realized branch. Stored trajectories retain at least
ten raw steps before teleportation to supply the training history.

We generate 25 additional branch pairs for validation. Each pair
shares the initial state, target, and pre-teleport actions, with mirrored
expert continuations and equal episode lengths. Targets have $y=112$ and
$x\in[180,202]$. Centering the target height makes the two continuations
mirror-symmetric. The generator verifies equal lengths and identical
pre-teleport states. Since evaluation windows end at episode completion,
unequal episode lengths could give the two branches different starting
images. Equal lengths keep the window starts at the same pre-teleport
state, while canonical goal rendering removes branch-dependent endpoint
offsets. Together, these checks ensure that both branches have identical
planner inputs.
Episodes contain at least 25 transitions and reach the
target within 20 transitions after teleportation. The final 25-transition
window therefore begins before the stochastic event. Both branches use a
canonical goal image with the agent exactly at the target. The planner
receives this image and the window's initial observation. Branch identity
is fixed by the episode and hidden from the model. Success uses a
16-pixel distance threshold. A predictor of the realized branch from
the shared initial and goal images has chance accuracy 50\% on these
balanced pairs.

\subsection{World-Model Training}
\label{app:planning-training}

We train each model on the corresponding dataset described above.
Appendix~\ref{app:model-training} gives the architecture, preprocessing,
and optimization settings for each model. Table~\ref{tab:training-settings}
lists the reporting epochs, observation noise, and objective weights.
We normalize continuous actions using statistics from the training trajectories.

\subsection{Planning and Evaluation}
\label{app:rollout-scoring}

\paragraph{Goal-conditioned planning}
We rank candidate action sequences by the mean squared distance between
their predicted terminal states and the encoded goal. \method uses all
image-patch coordinates. LeWM and SMWM use their
compact embeddings. CEM updates a Gaussian action distribution from the
best candidates using the settings in Table~\ref{tab:planning-settings}.

\begin{table}[htbp]
    \setlength{\belowcaptionskip}{6pt}
    \centering\small
    \caption{Shared CEM settings and evaluation budgets. A raw step is one environment
    transition, not one five-action prediction block.}
    \label{tab:planning-settings}
    \begin{tabular}{lcc}
        \toprule
        Setting & Four deterministic tasks & Teleport \\
        \midrule
        Candidates / elites / iterations & 300 / 30 / 30 & 300 / 30 / 30 \\
        Initial action variance (normalized) & 1 & 1 \\
        Fixed validation cases & 50 windows per task & 25 pairs / 50 windows \\
        Initial observations / past actions & 1 / none & 1 / none \\
        Goal offset (raw steps) & 25 & 25 \\
        Predicted blocks $H$ & 5 & 5 \\
        Actions per block $m$ & 5 & 5 \\
        Executed blocks before replanning & 5 & All 5, no replanning \\
        Total execution budget (raw steps) & 50 & 25 \\
        \bottomrule
    \end{tabular}
\end{table}

We allow at most two planning calls on deterministic tasks and execute
one plan without intermediate feedback on Teleport. We use simulator
states to restore initial conditions and determine success. The model
receives only images and action history.
Each task uses its native success criterion, including Teleport's
positional threshold described above.

\paragraph{Terminal conditional-mean scoring}
A sampled terminal state adds randomness to the goal cost even when the
preceding imagined history is fixed. The \emph{Final Mean} protocol reduces
this source of noise by scoring the final conditional mean. Intermediate
states are still sampled as in Equation~\ref{eq:latent-sampling}:
\begin{equation*}
    C(a_{\mathrm{traj}}) =
    \frac{1}{d}\left\|\mu_\phi\!\left(
    \hat{z}_{t+H-1},e_{a_{t+H-1}},\hat{h}_{t+H-1}
    \right)-z_g\right\|_2^2.
\end{equation*}
The hatted history contains observed context followed by sampled states.
We use one rollout per candidate and apply Final Mean to \method in
Figure~\ref{fig:planning-success}. The \emph{Sample} variant also samples
the final transition. Final Mean removes this last source of cost variation
while retaining uncertainty from earlier transitions. It omits the
variance term $d^{-1}\sum_j\sigma_j^2$ that would appear in the expected
squared distance conditioned on the same history.

\Needspace{12\baselineskip}
\subsection{Token-wise Prediction}
\label{app:tokenwise-prediction}

We also evaluate a token-wise autoregressive predictor on Teleport,
following TARFlow's autoregressive design~\citep{zhai2025normalizing}.
It retains a context window of three frames' worth of tokens and the
training settings of the parallel predictor. Conditioning changes from
frame-wise to token-wise: each next-frame patch depends on preceding
tokens and actions, including patches already generated within that frame.
Both predictors are autoregressive over time, while token-wise prediction
also models dependencies within each future frame.

\begin{figure}[htbp]
    \centering
    \includegraphics[width=\linewidth]{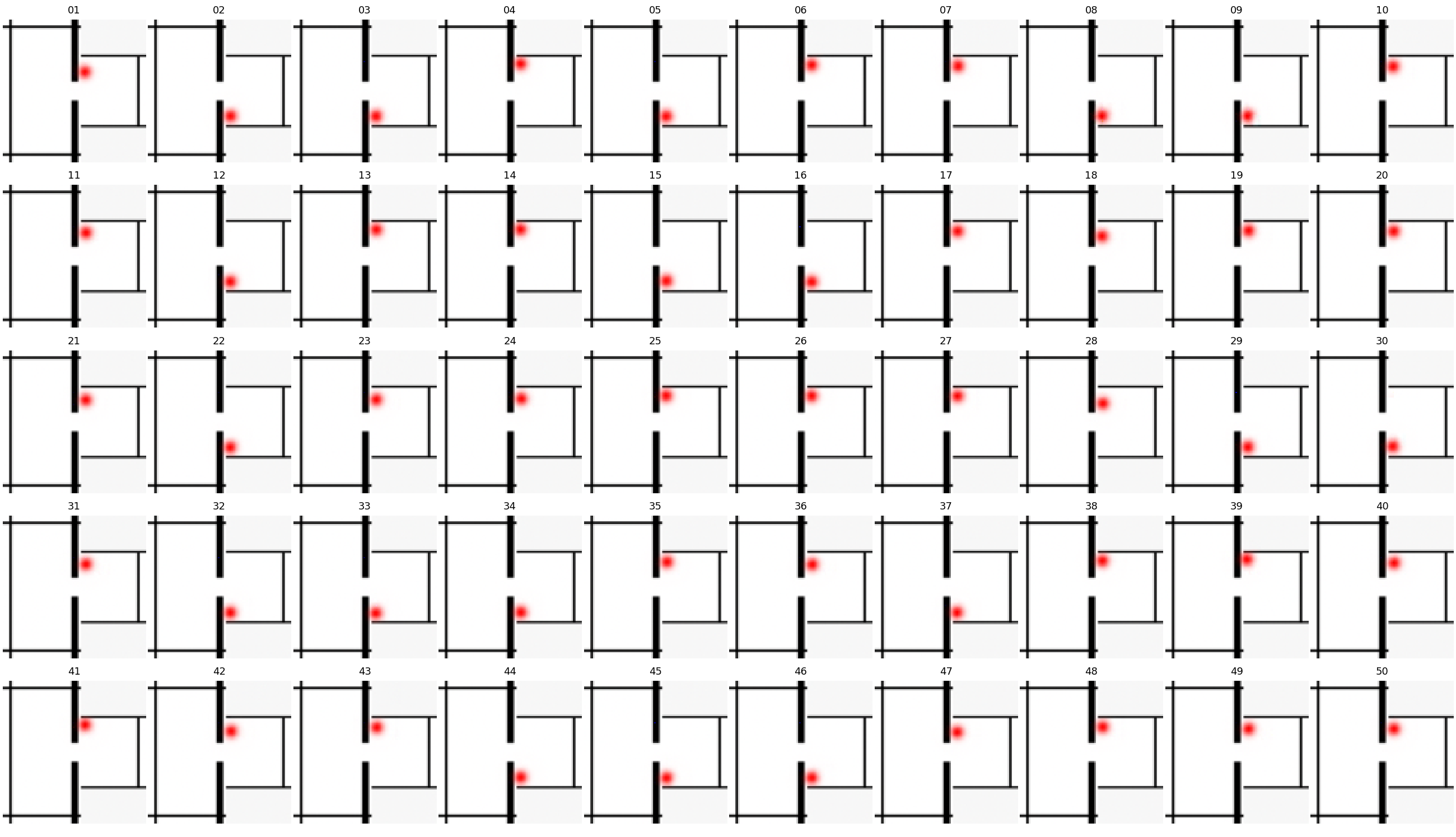}
    \caption{\textbf{Token-wise autoregressive prediction on Teleport.}
    Each patch is conditioned on previously generated patches within the
    same frame. Samples are decoded and score-denoised for display.}
    \label{fig:teleport-tokenwise}
\end{figure}

Figure~\ref{fig:teleport-tokenwise} shows decoded predictions, with score
denoising applied for display as described in Section~\ref{sec:generation}.
The token-wise model achieves 50\%
planning success in one run on 50 evaluation cases, using the same CEM
budget as the main experiment and sampled terminal states. Its sequential
patch generation increases planning time (Appendix~\ref{app:compute-followup}).

\clearpage
\section{Action Selection}
\label{app:libero-selection}

\subsection{Environment and Dataset}
\label{app:libero-data}

We use a deterministic custom version of the \emph{put the white mug on the plate} task
from LIBERO~\citep{liu2023libero}, through the LIBERO-PRO implementation~\citep{zhou2025liberopro}.
Initial object configurations are perturbed with magnitude $d=0.115$.
After reset and object placement, we advance the simulator for ten
zero-motion control steps with the gripper commanded open, allowing
contacts and object poses to settle before recording the initial state.
These initialization steps precede the evaluated action sequence.
We collect trajectories with the $\pi_{0.5}$ policy. The released training
set contains 18,685 episodes and 4,142,719 frames, with approximately
70\% successful and 30\% unsuccessful episodes by design. For action
selection, each world model receives images from one external RGB camera
and seven-dimensional actions describing translation, rotation, and gripper
control. The dataset also records wrist images and proprioception.
Appendix~\ref{app:libero-candidates} describes the shared evaluation pool.

\subsection{World-Model Training}
\label{app:libero-training}

Appendix~\ref{app:model-training} gives the architectures and optimization settings for the LIBERO models.
The LIBERO row of Table~\ref{tab:training-settings} specifies the reporting
epoch, observation noise, and inverse-dynamics weight.

\subsection{Candidate Construction}
\label{app:libero-candidates}

We construct a fixed pool of 50 scenes with 50 candidates per scene.
From each reset state, $\pi_{0.5}$ generates fresh action sequences and
may use closed-loop feedback during collection. We then freeze the
sequences and add independent Gaussian noise to the first six action
dimensions. Its standard deviation is half the corresponding
training-action standard deviation. We clip perturbed values to the
training-action extrema and leave the gripper command unchanged.
Incomplete final blocks are padded with zero motion and the last gripper
command for both scoring and execution.

We adjust the pool difficulty while retaining at least one candidate per
scene that succeeds in the environment, without inserting a ground-truth
action sequence. The pool contains 651 successful candidates out of
2,500, giving uniform random selection 26.04\% expected success and an
oracle 100\% success. These statistics characterize a success-conditioned
ranking task.
LeWM achieves 23.3\% success and SMWM achieves 32.7\%, compared with
46.0\% for \method. Candidate sequences, outcomes, and reset states are
shared across all models.

\subsection{Scoring and Execution}
\label{app:libero-scoring}

Each model receives the initial image and the terminal goal image from a
successful reference trajectory.
We predict the outcome of each candidate
using the action blocks and context settings in
Table~\ref{tab:shared-model-settings}, then execute the sequence with the
lowest terminal latent cost from the saved reset state. Execution uses no
replanning, and success follows the native LIBERO task criterion, checked
throughout the sequence. All models share the candidate pool and evaluation
randomness, with reporting checkpoints and observation noise specified in
Table~\ref{tab:training-settings}. \method uses a clean goal and one rollout
per candidate. Baseline inputs are clean.

\subsection{Candidate-Ranking Diagnosis}
\label{app:libero-diagnosis}

We evaluate how often the predicted ranking places a successful candidate
first or within its top five.
Table~\ref{tab:libero-ranking-diagnosis} reports the mean $\pm$ std of success
across three training runs of \method, LeWM, and SMWM at epoch 30, using the same frozen
candidate pool and evaluation randomness. We first rank candidates by
their predicted terminal costs. Top-1 measures the success of the selected
candidate. Top-5 measures whether any of the five highest-ranked candidates
succeeds. Uniform random ranking yields 70.30\% top-five coverage,
averaging the without-replacement probability across scenes.
Simulator Top-1 ranks the same candidates by the latent distance between
their encoded simulator terminal observations and the goal, then reports
the success of the lowest-cost candidate. This diagnostic evaluates
candidate ranking using the realized outcomes. For \method, endpoint
observation noise is shared across candidates and the goal is clean.
Success can occur at any step, whereas the score uses only the terminal
observation. Figure~\ref{fig:libero-top5} uses sample 4 and the first
\method training run at epoch 30. It is the first scene by sample ID with
a failed top-ranked candidate and at least one successful candidate in the
top five. The displayed images are actual simulator terminal observations.

\subsection{Image Reconstruction}
\label{app:libero-reconstruction}

\begin{figure}[htbp]
    \centering
    \includegraphics[width=\linewidth]{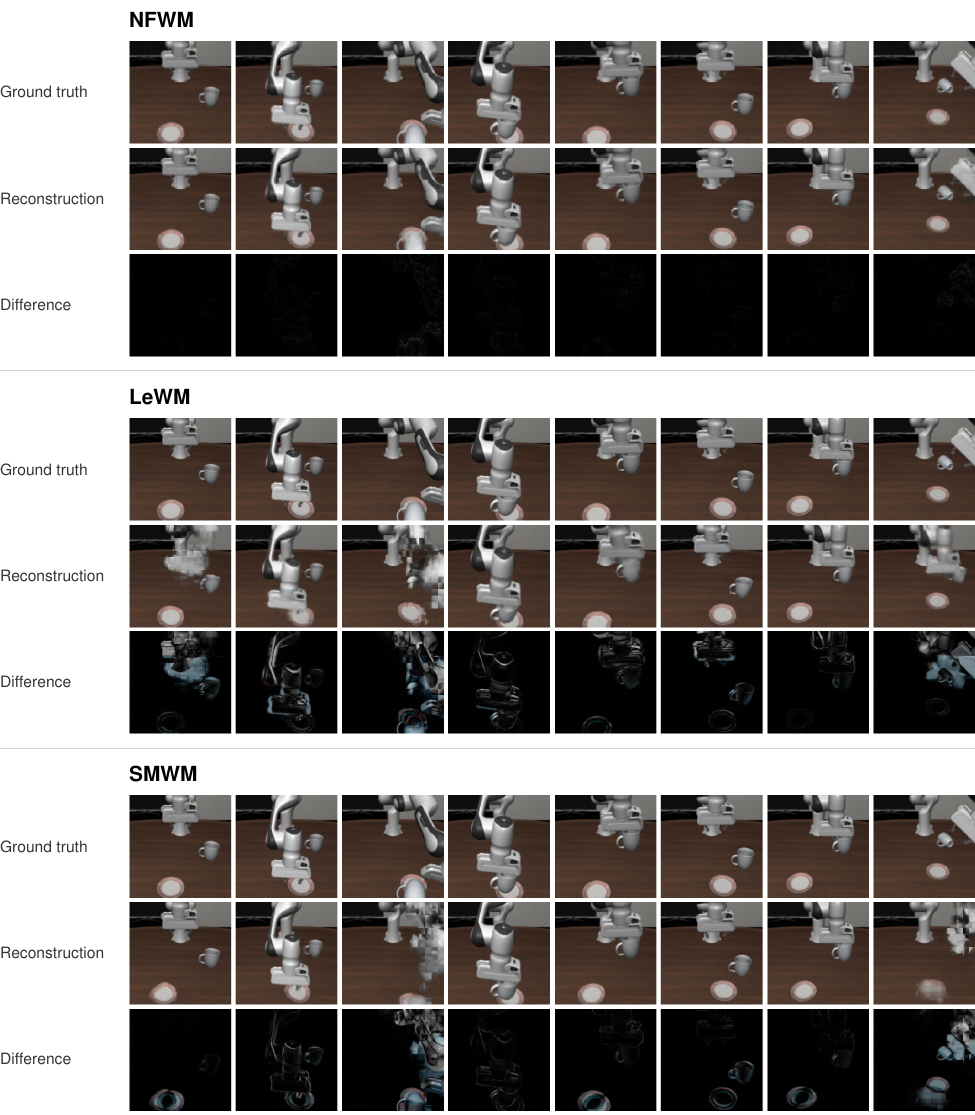}
    \caption{\textbf{Image reconstructions on LIBERO.}
    For each model, we show eight original observations, their reconstructions, and the corresponding difference images.
    LeWM and SMWM use 24-layer decoders.}
    \label{fig:libero-reconstruction}
\end{figure}

We reconstruct LIBERO observations from their encoded latent states to examine which visual details each representation preserves
(Figure~\ref{fig:libero-reconstruction}). We use Transformer decoders with architectures similar to each model's encoder and train them for 100 epochs with pixel MSE while keeping the encoders frozen. We use the same reconstruction protocol for \method and train a decoder on its frozen features. For LeWM and SMWM, the initial 12-layer decoders produce poor reconstructions, and increasing their depth to 24 layers still leaves visible errors, as shown in the figure.

\method more closely preserves the gripper, mug, and plate in these
examples. LeWM and SMWM recover the broad scene layout but distort or blur
parts of the robot and objects. These errors persist with deeper decoders,
indicating that fine visual details are less accessible in their learned
representations. Such details distinguish the small changes in gripper
pose and object alignment needed for LIBERO action selection.

\clearpage
\input{main/appendix_flappybird.tex}

\clearpage
\section{Policy Learning}
\label{app:policy-learning}

\subsection{Environment and Dataset}
\label{app:boxing-data}

We introduce stochastic opponent behavior into a two-player Atari Boxing
environment built on the Arcade Learning Environment
(ALE)~\citep{bellemare2013arcade}. A synchronized single-player ROM supplies
the opponent's actions. On each ALE frame, we independently replace its action with
probability 0.50, sampling uniformly from the other 17 actions.
The agent chooses among 18 actions and repeats
each choice for five ALE frames. Opponent randomness is applied throughout
this repetition. Sticky actions are disabled. Observations are
$96\times96$ RGB images, and unclipped player-one rewards are summed over
the repeated frames. Episodes terminate at ALE game-over, with no
time-limit truncations in the collected data.

We collect trajectories from ten checkpoints of one PPO run, spanning
roughly one to ten million training environment steps. Each checkpoint
generates 73 complete episodes and remains fixed within an episode.
We sample actions from temperature-scaled categorical logits. Stronger
checkpoints use higher temperatures to target returns near 30, while the
earliest low-return checkpoints use temperature one. We retain all
collected episodes regardless of return, yielding 1,029,628 transitions in
730 episodes with a mean return of 25.30 across the collection.

For each collector checkpoint, we assign 66 episodes to training and
seven to offline validation (Table~\ref{tab:boxing-data}). We split the
70 validation episodes equally between head selection and calibration
fitting (35 episodes) and a separate calibration check (35 episodes).
This partition is fixed across runs. Final policy evaluation uses
independent environment episodes, as described in
Appendix~\ref{app:atari-evaluation}.

\begin{table}[htbp]
    \setlength{\belowcaptionskip}{6pt}
    \centering\small
    \caption{Atari Boxing dataset. Counts refer to agent-decision
    transitions after five-frame action repetition, not raw ALE frames.}
    \label{tab:boxing-data}
    \begin{tabular}{lrrr}
        \toprule
        Partition & Episodes & Transitions & Mean episode return \\
        \midrule
        Training & 660 & 930,868 & 25.53 \\
        Offline validation & 70 & 98,760 & 23.19 \\
        Full collection & 730 & 1,029,628 & 25.30 \\
        \bottomrule
    \end{tabular}
\end{table}

\subsection{World-Model Training}
\label{app:boxing-model-training}

We train on consecutive stored transitions, each containing five ALE
frames and one categorical action, without an additional stride.
LeWM and SMWM use ViT-Tiny encoders with $8\times8$
patches on $96\times96$ images and ImageNet normalization. \method follows
Table~\ref{tab:nfwm-architecture}, with an 18-way inverse-dynamics head
trained by cross-entropy with weight 0.1. Gaussian observation noise has
standard deviation 0.05 for \method only. The same noisy encoding is used
for training the task heads, behavior cloning (BC), policy training, and environment evaluation.
LeWM and SMWM receive clean images. LeWM uses SIGReg weight 0.09,
and SMWM uses inverse-dynamics cross-entropy with weight 1. Reward and
termination heads are trained after freezing the world model.
Table~\ref{tab:boxing-training} specifies the budgets for these three models.

\begin{table}[htbp]
    \setlength{\belowcaptionskip}{6pt}
    \centering\small
    \caption{Atari Boxing settings for \method, LeWM, and SMWM, fixed across training
    repetitions except for the model-specific objectives and input noise
    described above. Selection follows Appendix~\ref{app:atari-evaluation}.}
    \label{tab:boxing-training}
    \begin{tabular}{p{0.40\linewidth}p{0.50\linewidth}}
        \toprule
        Setting & Value \\
        \midrule
        Images / action space & $96\times96$ RGB / 18 categorical actions \\
        World-model training & 50 epochs, learning rate $5\times10^{-5}$,
        shared settings in Table~\ref{tab:shared-model-settings} \\
        Frozen predictor-head training & 50 epochs, batch 128, AdamW,
        weight decay $10^{-3}$ \\
        BC budget & 30,000 updates from scratch \\
        BC / RL development evaluation & Every 1,000 updates, 30 episodes,
        sampled actions \\
        Imagination budget & 1,000 critic-only warm-up updates,
        then 20,000 actor updates \\
        Imagination horizon / batch & 20 agent steps (100 ALE frames) / 32 starts \\
        Actor--critic optimization & Learning rate $10^{-4}$, weight decay 0,
        gradient norm clip 1 \\
        Return / regularization & $\gamma=0.985$, $\lambda=0.95$,
        entropy coefficient $10^{-3}$, BC-reference KL coefficient 0.1,
        critic SmoothL1 weight 0.5 \\
        Final policy reporting & Final 20,000 actor updates, 50 held-out
        episodes with sampled actions \\
        \bottomrule
    \end{tabular}
\end{table}

\paragraph{JEDI adaptation}
\label{app:jedi}
Our offline JEDI~\citep{lim2026jedi} adaptation follows the Boxing dataset,
policy-learning budget, and evaluation protocol above.
Observations are resized to $64\times64$ with antialiased bilinear
interpolation and scaled to $[0,1]$. A residual CNN with channel widths
$32,32,32,16$ produces an $8\times8\times16$ latent. The width-160
convolutional U-Net concatenates three context latents and a noisy target,
with self-attention in its two middle residual blocks. Action and noise
embeddings condition the blocks through adaptive group normalization.
Encoder outputs and denoised latents use the soft clamp $3\tanh(x/3)$.
The encoder and denoiser have approximately 0.034M and 4.31M parameters.

We train for 50 epochs with batches of 64 clips, each containing three
context frames and three targets. The objective is preconditioned residual
MSE with stop-gradient latent targets. For each batch, subsequent history
uses encoded observations or denoised predictions with equal probability.
We draw noise levels with $\log\sigma\sim\mathcal{N}(-0.4,1.2^2)$, clip $\sigma$ to $[0.002,20]$, and use $\sigma_{\mathrm{data}}=1$. Inference uses three stochastic
Euler steps over $[0.002,5]$, with schedule exponent $\rho=7$, churn 1,
and unit-Gaussian initialization. There is no reconstruction loss,
inverse-dynamics loss, SIGReg, or EMA target encoder.

Separate reward and termination heads use an action-conditioned width-32
residual CNN, a 512-unit LSTM, and a 512-unit MLP. They predict three
reward-sign classes and two termination classes with unweighted
cross-entropy on detached encoded states. Each diffusion update alternates
with one head update on 32 clips. Separate AdamW optimizers use learning
rates $10^{-4}$ ($3\times10^{-5}$ for the encoder), weight decay $10^{-2}$,
and gradient-norm clipping at 1. Learning rates remain constant after 1,000 diffusion or 100 head warm-up updates. We use the final epoch-50 model and heads, calibrating termination before policy learning.

\paragraph{DreamerV3 adaptation}
\label{app:dreamer}
We adapt DreamerV3~\citep{hafner2023dreamerv3} to the same offline Boxing
setting. Its CNN encodes $96\times96$ observations into a recurrent
state-space model (RSSM), with a 2,688-dimensional deterministic state
and 32 categorical variables with 20 classes each. Reward and termination
heads are trained jointly with the world model.

To match the three-frame observation context, we rebuild the RSSM from
zero using only the latest three frames and actions at each decision.
During imagined rollout, the state is likewise rebuilt from the rolling
three-frame history. This restricts effective memory as well as the
training window. We train for 50 epochs on batches of 128 three-transition
clips, with learning rate $4\times10^{-5}$, 1,000 warm-up updates, and
bfloat16 mixed precision. We retain the final epoch-50 model for each
of three training seeds (3072, 3073, and 3074). Behavior cloning,
imagination training, and final evaluation follow the Boxing policy
protocol below. Table~\ref{tab:boxing-results} reports the final-policy
return of $64.47\pm6.32$.

\subsection{Policy Training and Evaluation}
\label{app:atari-evaluation}

We train a policy using experience generated by the frozen world model.
For \method, LeWM, and SMWM, we first fit reward and termination predictors.
JEDI and DreamerV3 use the heads trained alongside their world models
(Appendix~\ref{app:boxing-model-training}).
We initialize actors through behavior cloning and then optimize them on
imagined trajectories.

\paragraph{Reward and termination prediction}
For \method, LeWM, and SMWM, an MLP jointly predicts reward and a
termination logit from the dynamics predictor's final hidden representation, which already encodes
the history and current action. For \method, the head operates on each
image-patch token. We average reward predictions and termination logits
over tokens, then apply a sigmoid to the averaged logit. LeWM and SMWM
use their compact predictor features.

We train the head on detached offline-training features using the budget
in Table~\ref{tab:boxing-training}. Reward prediction uses SmoothL1 loss.
Termination prediction uses binary cross-entropy with positive examples
weighted by 1,000. The shared head is
LayerNorm--Linear(256)--GELU--Linear(2), with no separate encoder-reward
branch. Both targets describe the transition $(o_t,a_t,o_{t+1})$.
Termination denotes ALE game-over rather than truncation, and the world
model remains frozen throughout head training.

Termination estimates determine imagined continuation and therefore affect
the policy targets. We calibrate these estimates and select the checkpoint
with the lowest reward-validation loss among those passing the checks in
Table~\ref{tab:head-selection}. We then freeze the head and calibration
parameters. A final calibration check uses disjoint episodes.

\begin{table}[htbp]
    \setlength{\belowcaptionskip}{6pt}
    \centering\small
    \caption{Reward/termination head selection. Calibration uses
    $\hat p_t=\operatorname{sigmoid}(a\ell_t+b)$ with $a>0$.
    \method, LeWM, and SMWM use the same rule. Head and calibration parameters are
    frozen before policy improvement.}
    \label{tab:head-selection}
    \begin{tabular}{p{0.30\linewidth}p{0.63\linewidth}}
        \toprule
        Component & Setting \\
        \midrule
        Selection / check & 35 / 35 disjoint episodes \\
        Calibration validation & Two-fold episode-level cross-fitting,
        at least ten positive examples per fold \\
        Eligibility & Finite metrics, calibrated Brier score and unweighted
        binary NLL each $\leq 1.25$ times the constant-prevalence reference,
        mean predicted termination rate $0.25$--$4$ times the observed rate \\
        Selection and final fit & Lowest reward SmoothL1 among eligible
        checkpoints, refit calibration on selection episodes only \\
        Initialization & Analytical $-\log(1000)$ logit correction for
        the positive-class training weight \\
        \bottomrule
    \end{tabular}
\end{table}

\paragraph{Actor--critic architecture}
The actor--critic uses a separate attention network to process latent
history and intervening actions, with the context length in
Table~\ref{tab:shared-model-settings}. We project states to width 256.
Actions use
32-dimensional embeddings followed by projection. Token, time, and
source-type embeddings identify the inputs. Actor and critic queries attend
to this history through two cross-attention blocks with eight heads and
MLP width 1,024. Separate outputs produce 18 action logits and a scalar
value. Inputs are \method's full patch state, LeWM/SMWM's compact states,
JEDI's spatial states, or DreamerV3's RSSM states. These policy queries are separate from the auxiliary
inverse-dynamics readout.

\paragraph{Behavior cloning and selection}
We train the BC actor from scratch on the offline training split using
the budget and evaluation schedule in Table~\ref{tab:boxing-training}.
For each run, we select the checkpoint with the highest mean development
return, breaking ties by the earliest update. Selection is independent
across runs, without matching returns across models or filtering by
offline NLL. We freeze a copy of the selected actor as the BC reference
and begin imagination training with a critic-only warm-up.
Table~\ref{tab:boxing-per-seed} lists the selected updates.

\paragraph{Imagination learning}
For \method, LeWM, and SMWM, we initialize each rollout from one training
observation and grow the context as in Table~\ref{tab:shared-model-settings}.
JEDI and DreamerV3 initialize their dynamics from three observed frames.
All use the horizon in Table~\ref{tab:boxing-training}. The actor samples actions throughout the
rollout. For \method, sampled latent states enter the dynamics history,
while the actor and critic receive the corresponding conditional means $\mu$
after the initial observation. Using the mean removes transition-sampling
noise from the actor's input, which we expect to make policy optimization
more stable. This follows the same rationale as terminal conditional-mean
scoring in Appendix~\ref{app:rollout-scoring}.
The frozen BC reference uses the same
history as the actor. Reward and termination predictions condition on the
sampled dynamics history. LeWM and SMWM use their deterministic predictions
for both histories.

We optimize the actor with REINFORCE~\citep{williams1992reinforce} using a critic baseline and
$\lambda$-returns. We detach advantages and standardize them within each
batch. Calibrated termination probabilities provide continuation weights.
The loss also includes actor entropy, critic SmoothL1, and
$D_{\mathrm{KL}}(\pi(\cdot\mid h)\|\pi_{\mathrm{BC}}(\cdot\mid h))$,
with coefficients in Table~\ref{tab:boxing-training}. Gradients update only
the actor--critic. We report the policy after 20,000 actor updates.

\paragraph{Environment evaluation and independent training runs}
At each environment decision, the actor encodes the new observation and
samples an action using the real latent and action history. Evaluation
retains the training observation preprocessing,
five-frame action repetition, opponent randomness, and unclipped rewards.
Development and reporting use separate episode sets with sizes given in
Table~\ref{tab:boxing-training}. Environment and action-sampling randomness
are fixed across models and training runs. Execution is closed loop.

Each independent run retrains the world model, predictor head, BC actor,
and imagination policy with fresh training randomness. The dataset,
validation partition, evaluation randomness, and other settings remain
fixed. For \method, LeWM, SMWM, JEDI, and DreamerV3, Table~\ref{tab:boxing-results} reports
the mean and std of the three run-level mean returns at the final
20,000-update actor checkpoint. Thus, the error bars describe variation
across training runs. We use neither development-best RL checkpoints nor
held-out returns to select this endpoint. For Wan, we train one world model with jointly trained reward and termination heads and follow the same BC and imagination pipeline. The actor--critic reads Wan VAE latents, the reward and termination heads read the outputs of the 14th transformer block, and imagination rolls out Wan-generated latents with 1, 2, or 4 NFE. The actor is initialized from the final 30,000-update BC checkpoint, which achieves a mean held-out return of 43.1.

\begin{table}[htbp]
    \setlength{\belowcaptionskip}{6pt}
    \centering\small
    \caption{Atari Boxing results by training seed. BC and final RL
    are mean sampled returns over the same 50 held-out environment seeds. The BC
    checkpoint is selected on separate development episodes. Final RL
    always uses 20,000 actor updates after critic warm-up.}
    \label{tab:boxing-per-seed}
    \begin{tabular}{lrrrr}
        \toprule
        Model & Run & Selected BC update & BC & Final RL \\
        \midrule
        \method & 1 & 29,000 & 7.68 & 86.52 \\
        \method & 2 & 27,000 & 9.30 & 77.64 \\
        \method & 3 & 16,000 & 10.58 & 67.96 \\
        \midrule
        LeWM & 1 & 24,000 & 7.54 & $-54.80$ \\
        LeWM & 2 & 22,000 & 7.52 & $-42.30$ \\
        LeWM & 3 & 30,000 & 5.46 & $-8.46$ \\
        \midrule
        SMWM & 1 & 26,000 & 16.84 & 54.64 \\
        SMWM & 2 & 27,000 & 14.24 & 70.26 \\
        SMWM & 3 & 19,000 & 15.86 & $-2.56$ \\
        \bottomrule
    \end{tabular}
\end{table}

\method improves over its BC initialization in all three runs,
whereas LeWM degrades in all three. SMWM improves in two runs and degrades
in one, producing a positive aggregate but large training-seed variation.

\paragraph{Policy learning curves}
Figure~\ref{fig:boxing-policy-learning}(a) tracks development return from the
selected BC initialization through imagination training. At each point,
we average real-environment returns over three independent training runs,
using the schedule in Table~\ref{tab:boxing-training} and identical
environment seeds across updates and models. LeWM and SMWM
improve initially and then deteriorate, whereas \method continues to
improve. These development curves use separate episodes from the held-out
BC and final-policy returns in Table~\ref{tab:boxing-per-seed}. JEDI and
DreamerV3 improve early and retain these gains through the final update.
All five curves report means over three training runs.

\subsection{Input-Noise Control}
\label{app:boxing-noise}

We test whether observation noise explains \method's advantage by making
its representations more robust. We retrain both baselines with the same
noise level throughout training and evaluation. LeWM still deteriorates
during policy learning, while SMWM loses its gain
(Table~\ref{tab:boxing-input-noise}). Matched input noise therefore does
not reproduce \method's improvement.

We repeat the complete LeWM and SMWM pipelines with independent Gaussian
noise of standard deviation 0.05 in $[0,1]$ RGB units, without clipping.
We add noise before ImageNet normalization, or equivalently rescale it by
channel after normalization. Noise is applied to every observed encoder
input, including world-model contexts and targets, head training and
calibration, BC, imagination initialization, and environment evaluation.
The deterministic latent transitions receive no additional noise.

We retain the data splits, training randomness, architectures, and
objectives from reference run 1, with budgets from
Table~\ref{tab:boxing-training}. We reselect each head by minimum reward
loss among checkpoints passing termination checks and select BC actors
independently under the same budget. Each control uses one training run
and evaluates the final actor on the same held-out environment episodes.

\begin{table}[htbp]
    \setlength{\belowcaptionskip}{6pt}
    \centering\small
    \caption{\textbf{Boxing input-noise control (one training seed).}
    BC and final-RL columns are mean sampled held-out returns. Reward loss
    is the selected predictor head's SmoothL1 on the head-selection split.
    Clean-input baselines and \method use reference run 1.
    The two noise controls retrain each baseline with noisy encoder inputs.}
    \label{tab:boxing-input-noise}
    \begin{tabular}{lrrrr}
        \toprule
        Model & Pixel-noise $\sigma$ & BC & Final RL & Reward loss \\
        \midrule
        \method & 0.05 & 7.68 & 86.52 & 0.02423 \\
        LeWM & 0 & 7.54 & $-54.80$ & 0.02854 \\
        LeWM & 0.05 & 6.62 & $-45.38$ & 0.02604 \\
        SMWM & 0 & 16.84 & 54.64 & 0.02676 \\
        SMWM & 0.05 & 16.06 & 15.86 & 0.02655 \\
        \bottomrule
    \end{tabular}
\end{table}

Noise improves LeWM's final return relative to its clean-input run, but
the result remains far below BC. SMWM loses its imagination-learning gain,
while BC performance and head-selection reward loss change little.
For SMWM, the final 1,000-update averages of imagined per-step reward and
KL to BC rise from
0.135 to 0.362 and from 1.67 to 3.39, respectively, despite lower real
return. This pattern is consistent with model-biased policy optimization:
the policy moves further from BC toward trajectories assigned higher
imagined rewards but achieves lower returns in the environment.
The control applies observation noise during both training and evaluation.

\subsection{Latent-Distribution Drift and Effective Dimension}
\label{app:boxing-fd}

\paragraph{Open-loop evaluation with fixed contexts}
We use one epoch-50 checkpoint per model, with clean inputs for LeWM and
SMWM and observation noise for \method. For JEDI, we use the CNN trained with a three-frame history (seed 3072) and sample in three diffusion steps. DreamerV3 uses
its seed-3072 three-frame-memory model. We fix 50,000
training-reference states and 50,000 held-out origins, with 5,000 origins
per PPO collector checkpoint balanced across episodes. The splits contain
660 and 70 episodes, respectively. Each origin has three context frames
and a complete 20-step future within one episode. Models recursively
predict under the same recorded actions at $h\in\{0,1,2,5,10,15,20\}$,
without future observation feedback or termination censoring. Each step
is five ALE frames. \method uses one rollout per origin at prior scale
one, with cached observation noise shared across overlapping contexts.
FD is computed on latents before decoding or visualization denoising.

\paragraph{Vector and patch-level Fr\'echet distance}
We compute Fr\'echet distance (FD) directly between latent distributions~\citep{dowson1982frechet}.
For empirical Gaussian moments $(m,C)$ and $(m_R,C_R)$, the distance is
\begin{equation*}
    \operatorname{FD}(Q,R)=\|m-m_R\|_2^2+
    \operatorname{tr}\!\left(C+C_R-
    2(C_R^{1/2}CC_R^{1/2})^{1/2}\right).
\end{equation*}
Before computing these moments, each coordinate is standardized with its
training-reference mean and standard deviation. Variances are floored at
$\max(10^{-12},10^{-6}\bar v)$, where $\bar v$ is the reference's mean
coordinate variance, computed separately at each patch position.
LeWM and SMWM each use their full 192-dimensional vector and report
$\operatorname{FD}/192$. DreamerV3 uses its 3,328-dimensional RSSM state
and reports $\operatorname{FD}/3328$. Predicted and real-future states
are both rebuilt from rolling three-frame histories.

For spatial latent states, we compute FD at each fixed position across
samples, divide by the channel dimension $d$, and average over the $P$
positions:
\begin{equation*}
    D_{\mathrm{spatial}}(h)=\frac{1}{P}\sum_{p=1}^{P}
    \frac{\operatorname{FD}(Q_{h,p},R_p)}{d}.
\end{equation*}
Both \method and JEDI use this same definition and reference
standardization, with $(P,d)=(144,192)$ and $(64,16)$, respectively.
This statistic captures within-patch covariance and normalizes the spatial
and channel counts. Absolute FD values depend on the learned representation,
so we assess each curve's growth relative to its real-future control.

\paragraph{Observed drift and real-future control}
In Figure~\ref{fig:boxing-rollout-fd}(b), SMWM's vector FD is higher than
LeWM's at one and two steps but grows less at longer horizons. This panel
also shows DreamerV3's RSSM-vector FD, which rises rapidly over the first
five steps and then approaches a plateau. Panel (c) shows patch FD for
\method and JEDI. \method stays close to its real-future control, while JEDI and
DreamerV3 drift further from theirs. Table~\ref{tab:boxing-fd-controls}
reports the 20-step values. Panel (b) uses a linear axis and panel (c)
uses a logarithmic axis. Solid lines show predictions and dashed lines
show encoded real futures under the same recorded actions.

\begin{table}[htbp]
    \centering\small
    \caption{\textbf{Boxing rollout drift at 20 steps.} FD per dimension
    from one checkpoint per model, compared with encoded real futures
    against the same training reference.}
    \label{tab:boxing-fd-controls}
    \begin{tabular}{llrr}
        \toprule
        Model & Latent representation & Predicted & Real future \\
        \midrule
        \method & $144\times192$ spatial & 0.00256 & 0.00201 \\
        LeWM & 192D vector & 0.07654 & 0.00100 \\
        SMWM & 192D vector & 0.04925 & 0.00235 \\
        JEDI & $64\times16$ spatial & 1.45600 & 0.00251 \\
        DreamerV3 & 3,328D RSSM & 0.31581 & 0.01279 \\
        \bottomrule
    \end{tabular}
\end{table}

\paragraph{Effective latent dimension}
\label{app:effective-dimension}
For unstandardized 192-dimensional embeddings, we compute the participation
ratio $r_{\mathrm{eff}}=(\sum_i\lambda_i)^2/\sum_i\lambda_i^2$, where
$\lambda_i$ are training-reference covariance eigenvalues. It is 55.59 for
LeWM and 1.95 for SMWM. Explaining 95\% of the variance requires 54 principal components for LeWM and only three for SMWM. SMWM's concentration in fewer directions agrees
with its reported analysis~\citep{ivashkov2026smwm} and provides a possible
mechanism for limiting rollout error accumulation (Section~\ref{sec:boxing-analysis}).

\clearpage
\section{Shared Settings and Additional Analyses}
\label{app:experimental-details}

\subsection{World-Model Training and Baselines}
\label{app:model-training}

For goal-conditioned planning, LIBERO, and Atari Boxing, we train
\method, LeWM, and SMWM from scratch using the
common settings in Table~\ref{tab:shared-model-settings}. LeWM and SMWM
retain their respective architectures and objectives, and all models use
the same data splits on deterministic tasks. During training, predictions
condition on observed histories. During multi-step rollout, predicted
states enter the history recursively.

\begin{table}[htbp]
    \setlength{\belowcaptionskip}{6pt}
    \centering\small
    \caption{World-model training and context settings for
    \method, LeWM, and SMWM on goal-conditioned planning, LIBERO, and Atari Boxing.
    Diagnostic experiments specify any changes to rollout initialization.}
    \label{tab:shared-model-settings}
    \begin{tabular}{p{0.35\linewidth}p{0.55\linewidth}}
        \toprule
        Setting & Value \\
        \midrule
        Training batch & 128 windows \\
        Maximum observation context & Three frames \\
        Training window & Four observations: up to three context frames
        and one target \\
        Continuous-action transitions & Five raw steps between observations,
        concatenate the five intervening actions \\
        Continuous-action normalization & Statistics fitted on training
        trajectories for all reported evaluations \\
        Decision-rollout initialization & One observed frame, grow the
        predicted context to the maximum length \\
        World-model optimizer & AdamW, weight decay $10^{-3}$,
        gradient-norm clip 1, bfloat16 \\
        \bottomrule
    \end{tabular}
\end{table}

\paragraph{Architecture and preprocessing}
Table~\ref{tab:nfwm-architecture} summarizes \method's architecture.
Prediction and planning use the full image-patch latent state, while an
auxiliary readout supplies features for inverse-dynamics training. We take
the difference between normalized query outputs from consecutive observed
states to predict a detached action encoding. Gradients update both state
encodings and the readout. We average the continuous-action inverse loss
over target coordinates and normalize latent NLL and log-determinant by
image dimensionality, combining the latter two terms with equal weights
throughout training.

\begin{table}[htbp]
    \setlength{\belowcaptionskip}{6pt}
    \centering\small
    \caption{\method architecture for continuous-action tasks.}
    \label{tab:nfwm-architecture}
    \begin{tabular}{p{0.32\linewidth}p{0.58\linewidth}}
        \toprule
        Component & Setting \\
        \midrule
        Image / latent & $96\times96$ RGB, $8\times8$ patches,
        144 tokens of width 192 \\
        Flow & Four blocks, three Transformer layers per block,
        three heads of width 64, MLP width 768, alternating token order \\
        Affine scale & Soft clipping at 4, positive softplus scale,
        lower bound 0.1 \\
        Latent predictor & Six layers, token width 192, 16 heads of
        width 64, MLP width 2048, dropout 0.1, action AdaLN \\
        Conditional variance & Learned diagonal standard deviation,
        lower bound $10^{-4}$ \\
        Auxiliary readout & Eight queries, two self-attention layers,
        three heads of width 64, MLP width 768, no dropout \\
        Readout input/output & Learned slot and query embeddings,
        1,536-dimensional flattened queries with non-affine LayerNorm \\
        IDM head & MLP hidden width 128, difference of readouts,
        detached encoded-action target of width 192 \\
        \bottomrule
    \end{tabular}
\end{table}

For continuous-action tasks, \method resizes images to $96\times96$,
scales them to $[0,1]$, and adds Gaussian noise to training contexts and
targets. At evaluation, observed context images receive this noise but
the goal image is clean. The flow has no auxiliary register coordinates.
The eight learned readout queries receive no noise and are excluded from
the predictive state and planning cost.
Scoring uses the latent predictions directly. LeWM and SMWM use clean
$224\times224$ images, ImageNet normalization, and ViT-Tiny encoders with
$14\times14$ patches. Resizing is bilinear with antialiasing and no random
cropping. Their six-layer predictors operate on compact frame embeddings.

\paragraph{Optimization}
For continuous-action tasks, \method and LeWM use learning rate
$5\times10^{-5}$, and SMWM uses $10^{-4}$. Table~\ref{tab:training-settings}
gives \method's task settings. LeWM uses SIGReg weight 0.09. SMWM predicts
normalized raw action blocks with inverse-loss weights 30, 5, 1, 0.1, 0.1,
and 1 for Push-T, Reacher, Cube, TwoRoom, Teleport, and LIBERO,
respectively.
The four deterministic-task inverse weights are those reported by
SMWM~\citep{ivashkov2026smwm}. The SIGReg coefficient follows the published
baseline setting~\citep{maes2026leworldmodel,ivashkov2026smwm}.
For Teleport and LIBERO, which extend beyond those papers' control tasks,
we use the reproduction settings listed above. Across tasks, we standardize
batch size and context length using Table~\ref{tab:shared-model-settings}.

\begin{table}[htbp]
    \setlength{\belowcaptionskip}{6pt}
    \centering\small
    \caption{Continuous-action training settings.
    Reporting epochs are shared by \method, LeWM, and SMWM. Noise
    standard deviations (in $[0,1]$ pixel units) and inverse-dynamics
    weights are for \method.}
    \label{tab:training-settings}
    \begin{tabular}{lrrr}
        \toprule
        Task & Reporting epoch & Noise std. & $\lambda_{\mathrm{inv}}$ \\
        \midrule
        Push-T & 10 & 0.05 & 0.03 \\
        Cube & 10 & 0.05 & 10 \\
        Reacher & 10 & 0.05 & 20 \\
        TwoRoom & 10 & 0.05 & 0.1 \\
        Teleport & 150 & 0.05 & 1 \\
        LIBERO & 30 & 0.08 & 10 \\
        \bottomrule
    \end{tabular}
\end{table}

\subsection{Video World Model Reference (Wan)}
\label{app:wan}

We fine-tune the frame-wise autoregressive Wan2.1-T2V-1.3B model from Causal Forcing~\citep{wan2025wan,zhu2026causal} as an action-conditioned video world model. It predicts Wan VAE latents of $224\times224$ frames, and each latent corresponds to one five-action block (four actions on Boxing). The text input is a fixed task prompt. We encode actions with a randomly initialized encoder that follows the LeWM action-encoder architecture, and each transformer block adds a zero-initialized AdaLN projection of the action embedding to its timestep modulation. We fine-tune all parameters with flow matching and teacher forcing for 100k iterations (250k on LIBERO) without EMA, using the settings in Table~\ref{tab:wan-settings}.

At inference, the model generates each latent with 1, 2, or 4 Euler steps, conditioning on the previously generated latents. For planning and LIBERO, we score candidates by the distance between the predicted final latent and the goal latent in VAE latent space. Planning uses the CEM settings in Table~\ref{tab:planning-settings}, and LIBERO uses the candidate pool in Appendix~\ref{app:libero-selection}.

\begin{table}[htbp]
    \setlength{\belowcaptionskip}{6pt}
    \centering\small
    \caption{Wan fine-tuning settings.}
    \label{tab:wan-settings}
    \begin{tabular}{p{0.35\linewidth}p{0.55\linewidth}}
        \toprule
        Setting & Value \\
        \midrule
        Initialization & Frame-wise autoregressive Wan2.1-T2V-1.3B \\
        Action conditioning & Per-block zero-initialized AdaLN \\
        Objective & Flow matching with teacher forcing \\
        Batch / iterations & 16 clips / 100k (250k on LIBERO), no EMA \\
        Optimizer & AdamW, learning rate $10^{-5}$, weight decay $10^{-2}$, gradient-norm clip 1, bfloat16 \\
        Sampling & Euler, 1, 2, or 4 NFE per latent \\
        \bottomrule
    \end{tabular}
\end{table}

\subsection{Objective Ablations}
\label{app:ablation-details}

We ablate the training objectives to assess their contributions to Push-T
planning (Table~\ref{tab:objective-ablations}). The full model minimizes
$\mathcal L_{\mathrm{nll}}+\lambda_{\mathrm{inv}}\mathcal L_{\mathrm{inv}}$.
The NF-only variant (w/o InvDyn) sets $\lambda_{\mathrm{inv}}=0$, retaining observation
likelihood training, learned variance, and the unchanged patch-state
architecture. The unused auxiliary reader is retained but receives no IDM
gradient and is excluded from planning costs.

The clean-input forward-MSE variant (w/o NLL) replaces both the Gaussian latent NLL and
the log-determinant loss with forward MSE:
\begin{equation*}
    \mathcal L_{\mathrm{MSE}} = \mathbb E\!\left[
    d^{-1}\|z_{t+1}-\mu_\phi(z_t,e_{a_t},h_t)\|_2^2\right],
\end{equation*}
where $z_t=\mathrm{Enc}_\theta(o_t)$ contains all image-patch tokens. Its objective is
$\mathcal L_{\mathrm{MSE}}+\lambda_{\mathrm{inv}}\mathcal L_{\mathrm{inv}}$.
The variance head is absent, input noise is zero, and rollout always uses
the deterministic prediction. Gradients pass through the future latent
target. IDM retains the two-layer query reader and detached encoded-action
targets, as in the full model.

The first three rows of Table~\ref{tab:objective-ablations} use the same held-out Push-T split and the shared settings
in Table~\ref{tab:shared-model-settings}. We evaluate each variant at the
reporting epoch in Table~\ref{tab:training-settings}.
The full and forward-MSE variants use inverse weight 0.1.
Full and NF-only training retain the main model's image noise and learned
variance. All three use the CEM settings and fixed cases in
Table~\ref{tab:planning-settings}, with fixed evaluation randomness and
all-patch MSE cost. In this ablation, both the initial and goal images are noisy for the
two likelihood-trained runs. The clean-input forward-MSE variant uses clean initial and goal images.

Table~\ref{tab:objective-ablations} uses Final Mean for the likelihood-trained
models and deterministic rollout for the forward-MSE variant. Entirely
sampled rollout gives 84\% for the full model and 34\% for NF-only.
Removing inverse dynamics leaves the other settings unchanged, while the
clean-input forward-MSE variant also changes observation noise and rollout sampling.
This ablation uses inverse weight 0.1 and noisy goals for the
likelihood-trained models. The main planning result uses weight 0.03 and
a clean goal. With these settings and a single training run, the
full-model entry in Table~\ref{tab:objective-ablations} differs from the
\method result in Figure~\ref{fig:planning-success}.

\paragraph{Noisy-input forward-MSE control}
We additionally train the forward-MSE-plus-IDM variant with input-noise
standard deviation 0.05 and inverse weight 0.1, retaining the full patch
state. It uses the same architecture and optimization settings as the
clean-input MSE variant, with no learned variance and deterministic
rollout at every step. We evaluate the epoch-10 checkpoint from training
seed 3072 on 50 Push-T cases with evaluation seed 100045, a clean goal,
and the CEM budget in Table~\ref{tab:planning-settings}. Success increases
from 28\% to 40\%, below the full model's 86\%. The input noise improves planning success under the forward-MSE objective. The likelihood objective also trains the predictive variance. Together with the inverse-dynamics ablation, these results
support both objectives for Euclidean goal-distance planning.

\subsection{Representation Analysis}
\label{app:representation-analysis}

\begin{table}[t]
    \setlength{\belowcaptionskip}{6pt}
    \centering
    \small
    \caption{\textbf{Additional Push-T representation probes.} Held-out $R^2$ for
    the full reference model and its NF-only ablation at the fixed reporting
    epoch.}
    \label{tab:representation-probes-full}
    \begin{tabular}{lrrr}
        \toprule
        Model & Action $R^2$ & State $R^2$ & Displacement $R^2$ \\
        \midrule
        \method & \textbf{0.669} & \textbf{0.619} & \textbf{0.354} \\
        NF only & $-0.003$ & 0.301 & $-0.006$ \\
        \bottomrule
    \end{tabular}
\end{table}

We examine how inverse dynamics changes latent geometry using PCA,
held-out linear probes, and local action interventions. We freeze the
Push-T models at the objective ablation's reporting epoch and analyze their
flattened image-patch latents. Auxiliary query outputs are excluded, since
they are untrained in the NF-only variant. All variants use identical
physical states and actions. The full model and NF-only share pixel-noise
draws, while the forward-MSE variant uses clean inputs. These diagnostics
use one checkpoint per variant. PCA illustrates the geometry. Linear
probes and local fits quantify it in the full latent space.

\begin{figure}[htbp]
    \centering
    \includegraphics[width=\linewidth]{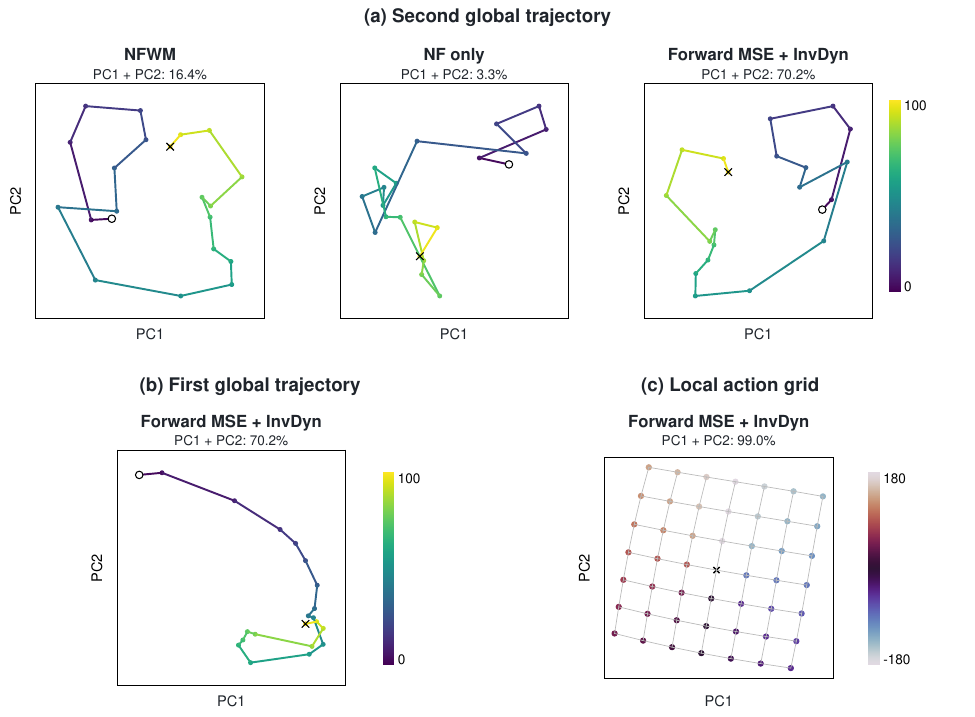}
    \caption{\textbf{Additional Push-T projections.}
    (a) A second real trajectory across all three objectives.
    (b) The first trajectory under forward MSE + IDM, complementing
    Figure~\ref{fig:geometry-main}(a). Color denotes elapsed time.
    (c) Local predictions for forward MSE + IDM under the same action
    grid as Figure~\ref{fig:geometry-main}(b). Color denotes action direction.
    The deterministic variant has orderly projections despite weaker planning.}
    \label{fig:global-pca}
    \label{fig:local-pca}
\end{figure}

\paragraph{Global trajectory PCA}
For each model, we encode six held-out trajectories of 21 observations
spaced five raw steps apart, center the pooled 126 states, and project onto
the first two principal components. Figure~\ref{fig:geometry-main}(a) shows
the first trajectory for the full model and NF-only.
Figure~\ref{fig:global-pca} adds the second trajectory and the forward-MSE
variant. Both plots show encoded real observations in their original
temporal order. The first two components explain 16.4\%, 3.3\%, and 70.2\% of
variance for the full, NF-only, and forward-MSE models, respectively.
IDM-trained representations show more temporally coherent paths in these
examples, whereas NF-only has more abrupt turns and overlapping segments.
For NF-only, the projection captures only a small fraction of total
variance, motivating the full-dimensional probes below.

\paragraph{Local predicted-state PCA}
For a fixed held-out state, we add a $7\times7$ grid of normalized
two-dimensional offsets in $[-0.3,0.3]^2$ to the same five-action expert
block. Each offset is applied to all five actions, with native environment
bounds enforced for this diagnostic. Each model predicts one block from
the same initial observation for all 49 candidates. We center the predicted
states and project onto their first two principal components. In this
diagnostic, the full model and NF-only share a latent noise draw across
candidates to isolate action effects. The forward-MSE variant is deterministic.
Figure~\ref{fig:geometry-main}(b) shows the full model and NF-only.
Figure~\ref{fig:local-pca}(c) adds the forward-MSE variant. The latter has a
regular projection despite weaker planning. We therefore also measure
action effects using encoded outcomes from the actual environment.

\begin{figure}[htbp]
    \centering
    \includegraphics[width=0.92\linewidth]{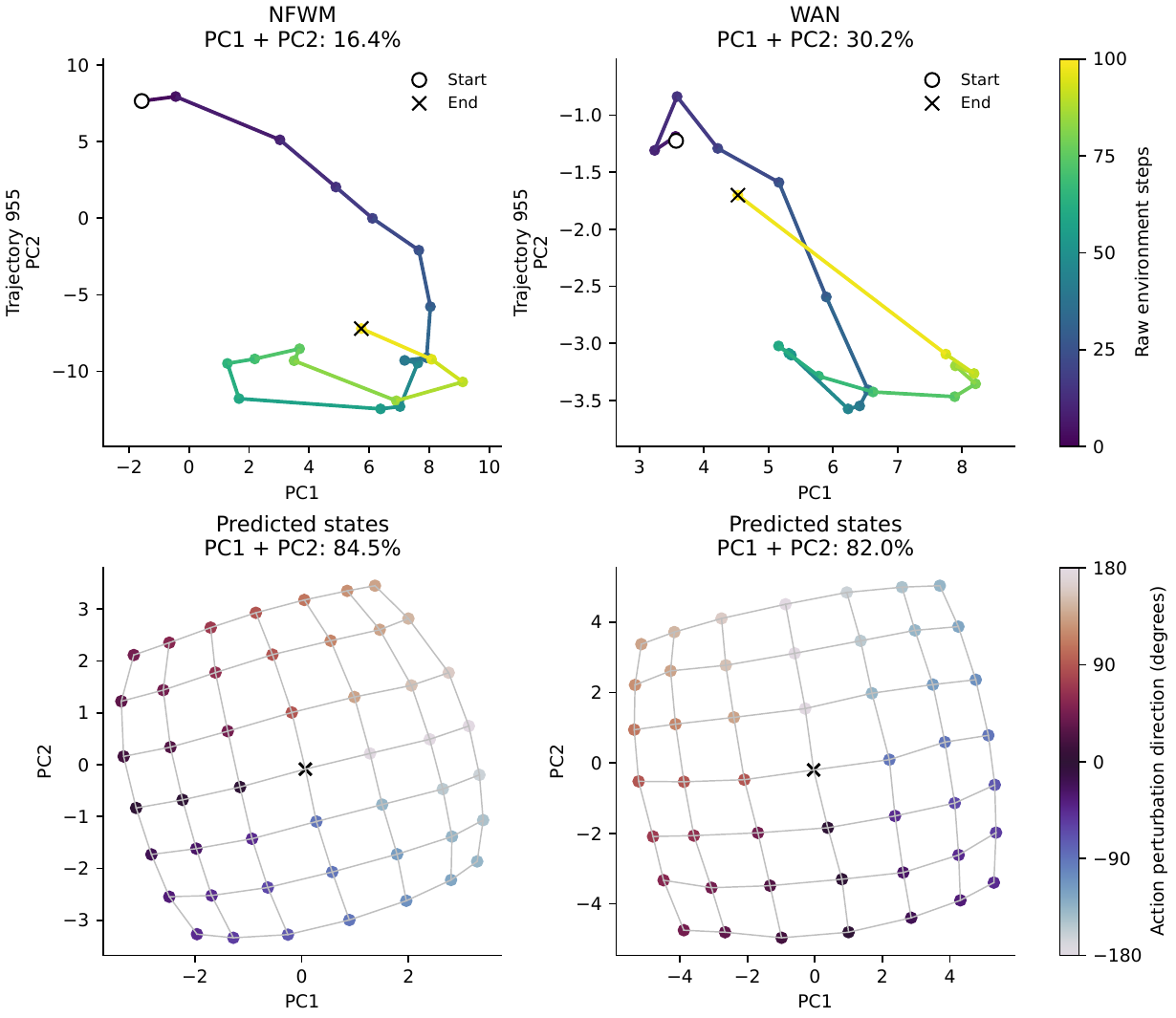}
    \caption{\textbf{Matched Push-T geometry of \method and Wan (1 NFE).}
    Top: encoded observations along the same real trajectory, colored by
    time. Bottom: predicted latent states for the same local action grid,
    colored by action direction. Wan shows sharper trajectory reversals,
    while both predictors form ordered local grids.}
    \label{fig:wan-vae-pca}
\end{figure}

\begin{figure}[htbp]
    \centering
    \begin{minipage}[t]{0.46\linewidth}
        \centering
        {\sffamily\bfseries\footnotesize Wan (2 NFE)}\par\vspace{2pt}
        \includegraphics[width=\linewidth,trim=270bp 0 0 269.1bp,clip]{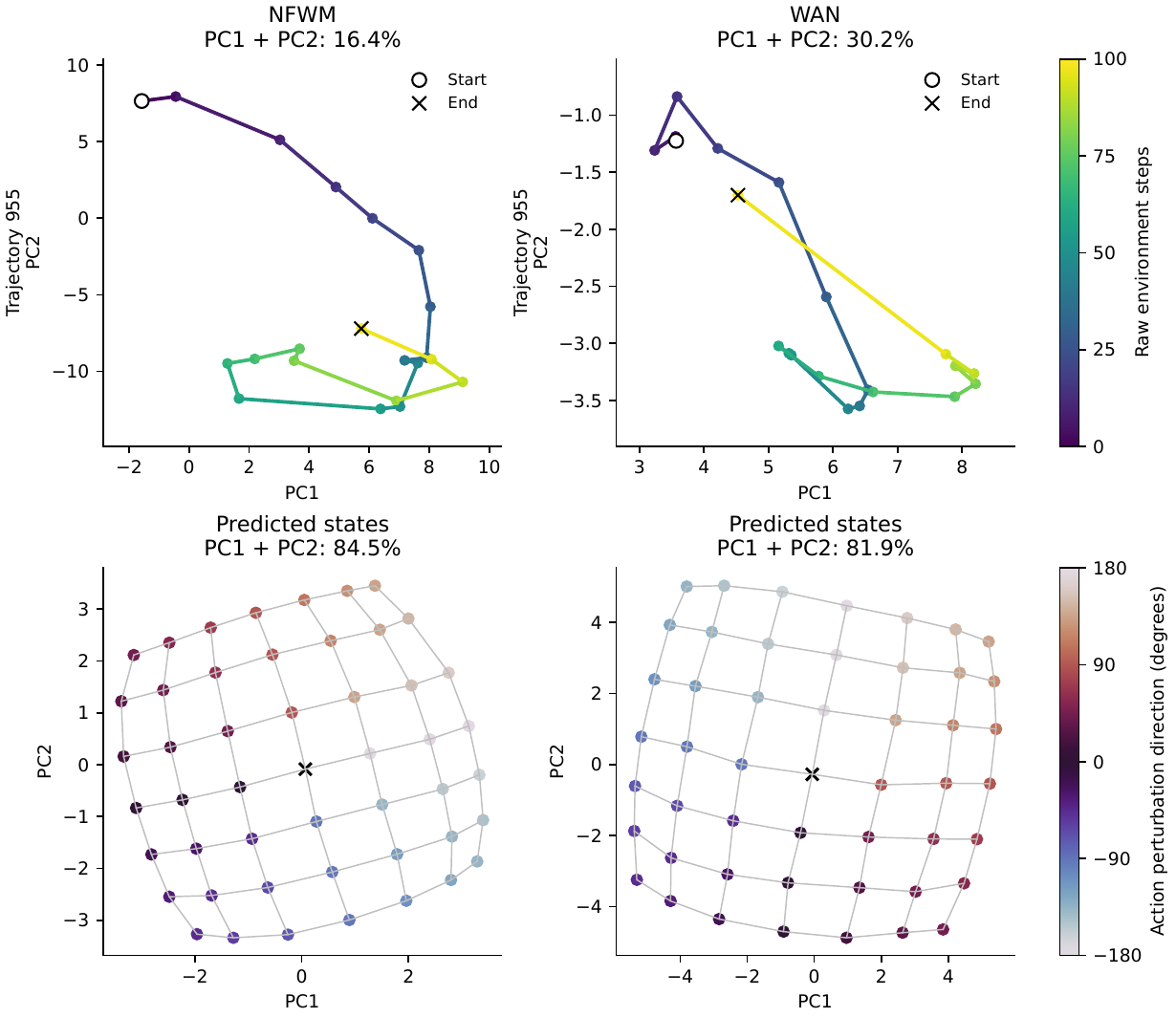}
    \end{minipage}\hspace{0.02\linewidth}%
    \begin{minipage}[t]{0.46\linewidth}
        \centering
        {\sffamily\bfseries\footnotesize Wan (4 NFE)}\par\vspace{2pt}
        \includegraphics[width=\linewidth,trim=270bp 0 0 269.1bp,clip]{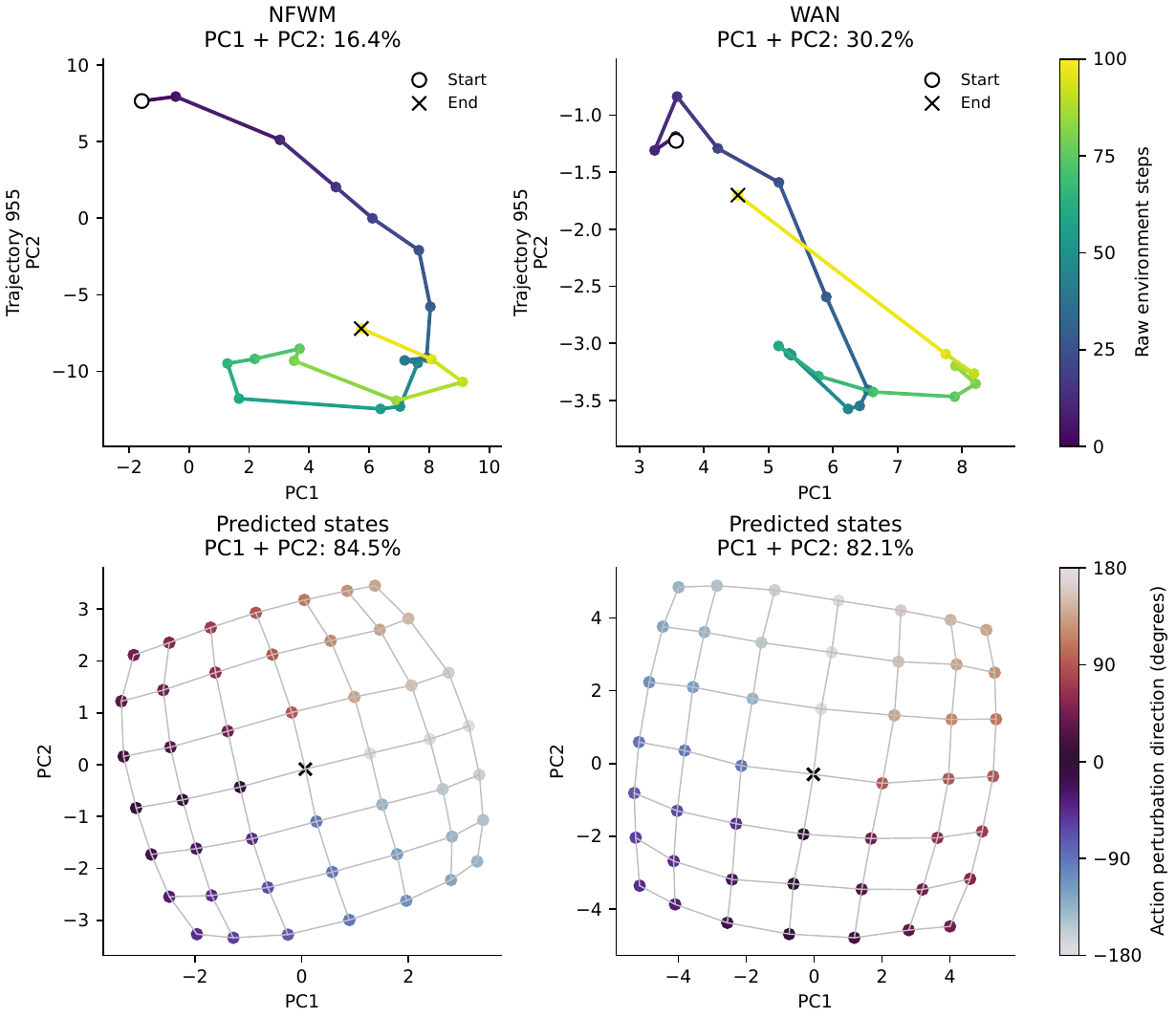}
    \end{minipage}
    \caption{\textbf{Local predicted-state geometry of Wan with 2 and 4 NFE.}
    The initial observation, action grid, and PCA protocol follow
    Figure~\ref{fig:wan-vae-pca}. The local grid retains a similar structure
    across the two sampling settings.}
    \label{fig:wan-pca-nfe24}
\end{figure}

\paragraph{Matched NFWM and Wan geometry}
We compare \method and Wan under the same PCA protocol, with matched
observations and action candidates. Global PCA uses encoded real observations
and does not depend on NFE, so Figure~\ref{fig:wan-vae-pca} shows it once.
Figure~\ref{fig:wan-pca-nfe24} compares Wan's local predictions with 2 and
4 NFE. Sampling noise is fixed across action candidates.

Wan's trajectory shows sharper reversals and tightly grouped states
connected by longer segments. Together with its lower planning success,
this suggests a geometry less suited to Euclidean-distance goal comparison.
Both local grids remain ordered, indicating coherent responses to small
action perturbations. Wan's local grid remains similar across 1, 2, and 4 NFE.

\paragraph{Held-out linear probes}
We fit ridge probes for the full model and NF-only on 800 distinct
world-model-training episodes. We use 200 additional training episodes to
select regularization and 400 held-out episodes for evaluation, sampling
one five-step transition per episode. We fit centering, feature scaling,
and target normalization on the probe-training split.
The regularization grid is $10^{-6},\ldots,10^2$. State targets comprise
agent and block positions and the sine/cosine of block orientation.
Displacement targets comprise their translations and wrapped block
rotation. The action target is the normalized raw five-action block.
State probes read $z_t$. Displacement and action probes read
$z_{t+1}-z_t$.
Table~\ref{tab:representation-probes-full} reports test $R^2$, averaged
uniformly over target coordinates. The full model scores higher on all
three probes, with particularly large gains in action and displacement
recovery. Inverse dynamics therefore makes control-relevant information
more accessible to a linear readout. These probes measure the accessibility
of information retained by both invertible encoders.

\paragraph{Quantifying local linearity}
We also reset the actual environment to each of 16 fixed held-out states
and execute the action grid at normalized radii 0.1 and 0.3. We fit an
affine map from the two-dimensional action offset to the full encoded
endpoint displacement on 30 actions and test it on 10 different actions.
The remaining nine actions select regularization for a complementary
inverse ridge probe. Common pixel noise across endpoints controls nuisance
variation in the representation comparison. The forward score is
$1-\mathrm{SSE}/\mathrm{SST}$, where SST uses the fit-action mean, averaged
over the 16 states. At radius 0.3, the full model scores 0.696 versus 0.497
for NF-only. At radius 0.1, the scores are 0.797 and 0.690.
Within these tested states and perturbations, inverse dynamics makes the
encoded effects of actions more closely follow an affine map.

The reverse local probe, from actual latent displacement to action offset,
is near ceiling for both models (0.997 versus 0.991 at radius 0.3).
This indicates that action effects remain recoverable without IDM.
The full model makes these effects more accessible to linear readout and
more regular in the forward direction. Combined with the forward-MSE
variant's orderly projections but weaker planning, this suggests that
useful geometry must also support accurate prediction and goal comparison.

\subsection{Decoding and Prediction Visualizations}
\label{app:rollout-visualization}

Figure~\ref{fig:decoded-rollouts} extends the main prediction examples over
longer rollouts. \method's Teleport samples retain separate destinations,
while the LeWM and SMWM point predictors initially merge them. Later LeWM and
SMWM Boxing frames show increasingly distorted player shapes, while
\method retains more distinct players in these examples. The decoded
sequences illustrate how rollout errors can alter the represented scene
structure, complementing the latent-distribution analysis in
Appendix~\ref{app:boxing-fd}.

\begin{figure}[htbp]
    \centering
\input{figures/decoded_rollouts.tex}
    \caption{\textbf{Decoded predictions in stochastic environments.}
    (a) Selected Teleport predictions at five-step intervals. The panels share the
    initial observation and first action block. Upper and lower \method
    continuations then follow branch-specific expert actions.
    (b) Atari Boxing predictions, progressing from left to right.
    Teleport and Atari Boxing \method images receive one score-denoising step for display.}
    \label{fig:decoded-rollouts}
\end{figure}

\paragraph{Decoding and display denoising}
For these visualizations, we train LeWM and SMWM decoders with pixel MSE
while keeping their encoders frozen. We use epoch-75 decoder checkpoints
for Teleport and epoch-200 checkpoints for Boxing.
\method maps sampled latents through the inverse flow. For display, we
apply one conditional score step following TARFlow~\citep{zhai2025normalizing}:
\begin{equation*}
    y_{\mathrm{vis}}=y+\sigma_{\mathrm{obs}}^2
    \nabla_y\log p_{\theta,\phi}(y\mid h,a),\qquad\sigma_{\mathrm{obs}}=0.05.
\end{equation*}
We hold the sampled history and action fixed and clip the resulting
values for display. The step uses no goal or future observation, and its
output is used only for visualization.

\subsection{Reporting Protocol and Statistical Details}
\label{app:result-reporting}

For goal-conditioned planning, LIBERO, and Atari Boxing, we evaluate independently trained \method, LeWM, and SMWM checkpoints on
the same cases or episodes, using the budgets in
Tables~\ref{tab:planning-settings} and~\ref{tab:boxing-training}. LIBERO
reuses the candidate pool in Appendix~\ref{app:libero-selection}.
Evaluation seeds fix CEM randomness and, for \method, observation noise
and latent sampling. Across training runs, we vary initialization,
shuffling, and training randomness while fixing the data split,
hyperparameters, reporting epoch, and evaluation cases. Wan results come from a single training run per task, so we report them without standard deviations.

For success percentages $r_1,\ldots,r_n$, we report
$\bar r=n^{-1}\sum_i r_i$ and sample standard deviation
$s=\sqrt{\sum_i(r_i-\bar r)^2/(n-1)}$ when $n>1$.

\FloatBarrier
\Needspace{8\baselineskip}
\input{main/appendix_followup.tex}

%% file: main/appendix_flappybird.tex
\section{Reward-Based Planning}
\label{app:flappy-bird}

\subsection{Environment and Dataset}

Flappy Bird provides a two-action visual control task: flap or no-op,
with $128\times72$ RGB observations. The dataset contains 8,289 episodes
and 1,315,748 frames. Of these episodes, 4,800 come from a scripted
controller and 3,489 are collision-focused examples, each with a 25-frame
segment ending in collision. Every episode appends 24 frozen post-collision
frames paired with random actions, zero reward, and termination labels of
one.

We split episodes 90/10 with seed 3072, using 7,460 for world-model
training and 829 for validation. Validation episodes are used only to
compute validation loss during world-model training. After freezing the
world model, we fit the readouts on all 8,289 episodes, including the 829
reserved for world-model validation. The separate validation and test
files, each containing 600 episodes, are not used for fitting either model.
Closed-loop evaluation uses live environment episodes as described below.

\subsection{World Model and Task Readouts}

\paragraph{Architecture}
The 5.9M-parameter TARFlow encoder uses $8\times8$ patches, producing
144 tokens of width 192. Four blocks of three causal Transformer layers
alternate direction along the patch sequence. The encoder is shared
across frames and receives no action or history. The 10.9M-parameter
predictor has six layers, model width 192, 16 attention heads of width 64,
MLP width 2,048, and dropout 0.1. It uses rotary temporal positions,
frame-causal attention over an eight-frame window, and a key--value cache.
Standardized actions are embedded to width 192 and condition every layer.
Two output projections predict the next frame's mean and scale, with
softplus and a $10^{-4}$ floor for the scale. The conditional density is
diagonal Gaussian over all $144\times192$ coordinates, as in
Section~\ref{sec:training-inference}.

\paragraph{World-model training}
We sample an episode uniformly, then a 25-frame clip within it, giving
24 prediction targets. Training uses 12 epochs of approximately one million
clips each, with batch size 64, AdamW learning rate $5\times10^{-5}$,
weight decay $10^{-3}$, gradient-norm clipping at 1, and bfloat16 precision.
We add Gaussian noise with standard deviation 0.05 to observations in
$[0,1]$ during training and validation. The objective in
Equation~\ref{eq:generative-objective} is normalized by the latent
dimension $144\times192$ and averaged over targets and clips. Inverse
dynamics, reward, and termination losses have zero weight. We use the
epoch-12 checkpoint for readout fitting and planning.

\paragraph{Readout architecture}
The readout uses the embedding output and all six predictor layers.
We apply nonparametric LayerNorm to each, average across layers, and
normalize again. These action-conditioned $144\times192$ features are
detached from the world model. After normalizing the flattened frame,
we add learned patch positions and two query tokens. Two Transformer
layers with three width-64 heads and MLP width 768 process the tokens.
Separate LayerNorm--MLP heads (192--256--1, GELU) map the two queries
to reward and a termination logit.

\paragraph{Readout fitting}
Reward targets are 0.1 for an ordinary surviving step, 1 for passing a pipe,
$-1$ for collision, and 0 for frozen frames. Termination targets are one
on collision and frozen transitions and zero otherwise. We fit the heads
with equally weighted SmoothL1 reward loss and unweighted binary
cross-entropy for termination. Training uses 3,000 updates, batches of
16 episode-sampled 25-frame clips, AdamW learning rate $3\times10^{-4}$,
weight decay $10^{-3}$, and gradient-norm clipping at 1. The world model
is frozen, and histories use clean observed frames through teacher forcing.

\subsection{Planning and Evaluation}
\label{app:flappy-planning}

\begin{figure}[htbp]
    \centering
    \includegraphics[width=\linewidth,trim=0 280bp 0 130bp,clip]{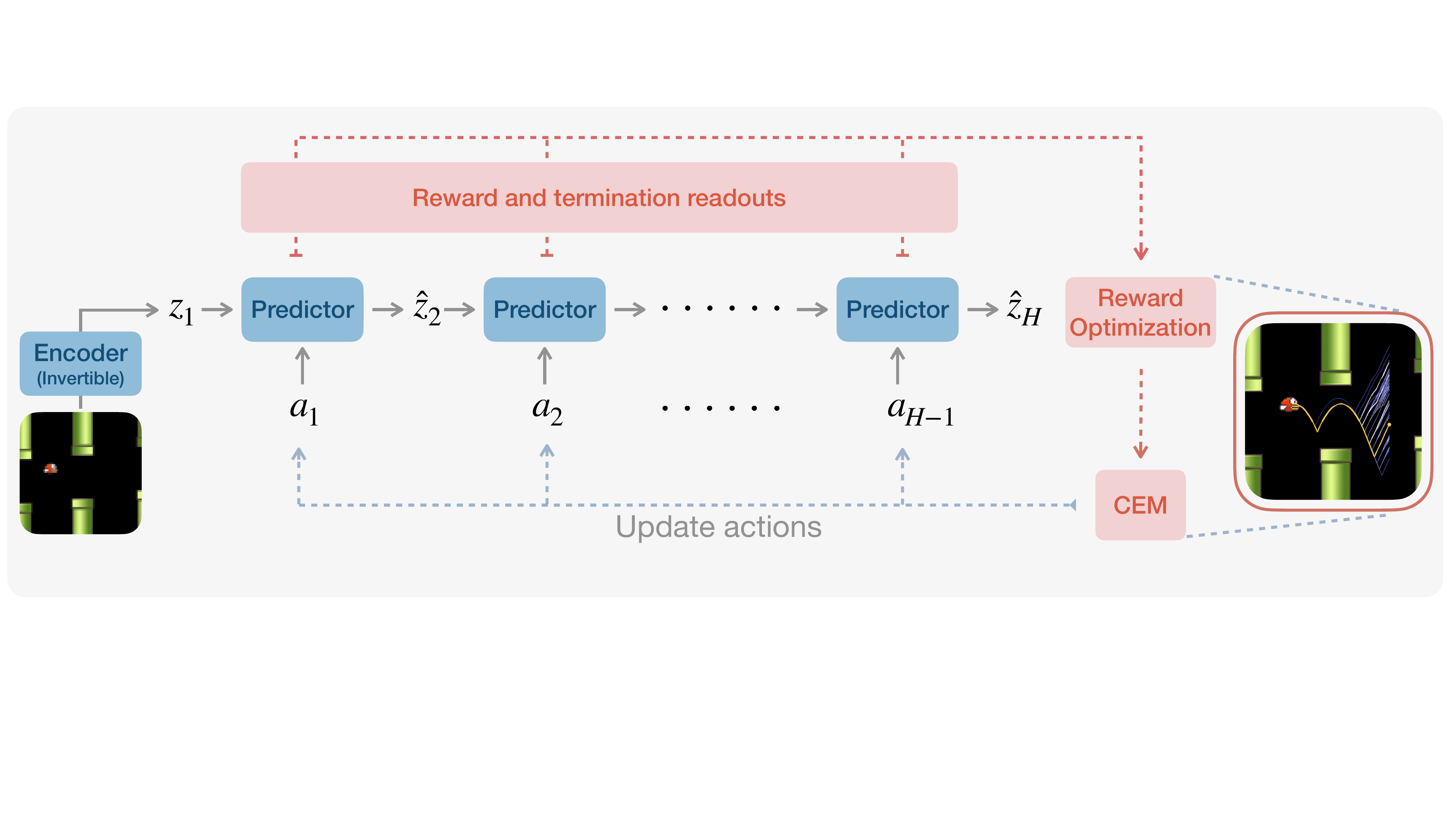}
    \caption{\textbf{Planning with \method in Flappy Bird.} The frozen predictor rolls each candidate action sequence forward. Its readouts estimate reward and termination at each step, and CEM uses the resulting scores to update the action distribution.}
    \label{fig:flappy-cem}
\end{figure}

Planning conditions on the latest eight observations and seven intervening
actions. Each candidate has 40 binary actions and one sampled latent
rollout (Figure~\ref{fig:flappy-cem}). At imagined step $k$, the frozen readouts predict death
probability $p_k$ and reward $\hat r_k$ from action-conditioned predictor
features. We clip the reward and accumulate survival-weighted scores:
\begin{equation}
    q_k=\operatorname{clip}(\hat r_k,0,1),\qquad
    S_k=\prod_{i=1}^{k}(1-p_i),\qquad
    J(c)=\sum_{k=1}^{40}S_k(1+q_k).
    \label{eq:flappy-score}
\end{equation}
The unit term rewards survival, while $q_k$ adds the clipped predicted
reward. A collision reduces subsequent contributions through $S_k$.
All latent transitions are sampled, including the final step.

\begin{figure}[htbp]
    \centering
    \begin{tabular}{@{}ccc@{}}
        \includegraphics[width=0.246\linewidth]{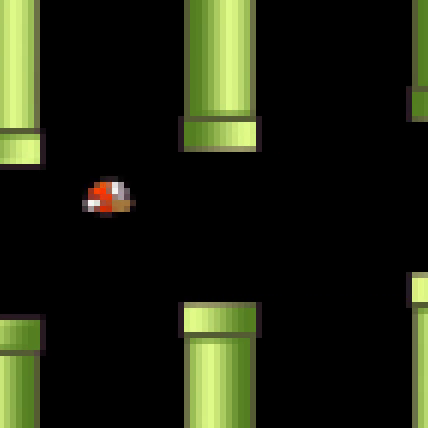} &
        \includegraphics[width=0.246\linewidth]{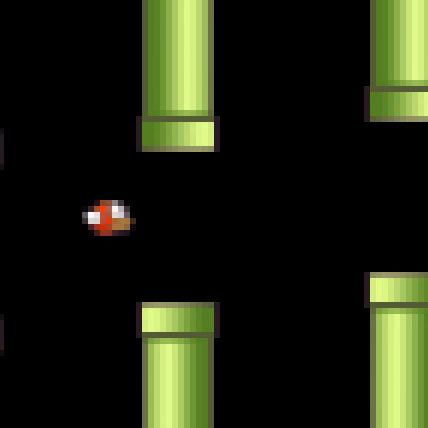} &
        \includegraphics[width=0.246\linewidth]{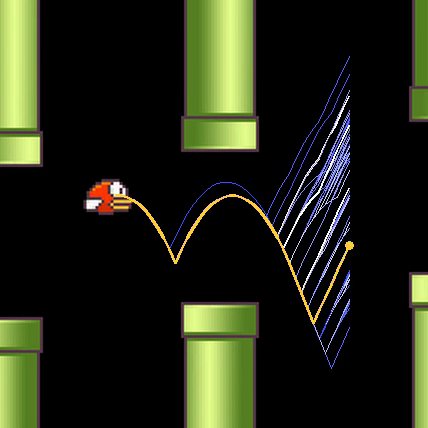} \\
        \small Observation at 225 & \small Observation at 232 & \small CEM at frame 225
    \end{tabular}
    \caption{\textbf{Observations and candidate plans in Flappy Bird.} The right panel shows candidate action sequences traced with simulator physics. Yellow marks the plan selected by CEM.}
    \label{fig:flappy-cem-example}
\end{figure}

CEM runs four rounds of 256 candidates, retaining the 64 with the highest
$J$. It updates each Bernoulli action probability to the corresponding
elite mean, clipped to $[0.02,0.98]$. In the first round, half the
candidates use the previous planning distribution shifted by one action,
and half use a flap prior of 0.1. We retain the shifted previous best
sequence and the current best sequence among the candidates. The agent
executes the first action of the best sequence across rounds, then observes
and replans (Figure~\ref{fig:flappy-cem-example}). Planning uses latent predictions and readouts without decoding.
A reference plan takes approximately 4.5 seconds on one H200.

Evaluation stops at death or 100 cleared pipes, with no frame cap. One
trained model is evaluated on level seeds $0,1,2,3,4,50,6,7,8,9$.
None of these level seeds is used for training. All ten games reach
100 pipes, each after 3,723 frames.

\paragraph{Action-consequence visualization}
Figure~\ref{fig:flappy-action-consequences} uses the eight-frame history
ending at frame 48 of level seed 0 and displays the fourth predicted and
executed step. Predictions are sampled without future observation feedback
and score-denoised for display (Section~\ref{sec:generation}). In the full
20-step example, the committed plan survives, never-flap actions lead to
collision at step 8, and always-flap actions collide at step 4. The
termination readout first exceeds 0.5 at steps 8 and 3 for the latter two
sequences, respectively.

\FloatBarrier
\subsection{Exploratory Deterministic Controls}

We train JEPA-style controls using latent MSE and SIGReg, following the
point-prediction formulation of LeWM~\citep{maes2026leworldmodel}. One uses
the invertible encoder, and another uses a non-invertible ViT with 5.9M
parameters. Their deterministic predictions are fed back during rollout.
They use the same training data, learning rate, batch size, detached
readout-training protocol, and seeded CEM budget as the likelihood-trained
model. SIGReg has weight 1, with no inverse-dynamics or reconstruction
loss. Each control uses one training seed without hyperparameter tuning.
These exploratory controls are separate from the LeWM and SMWM
reproductions used in the other benchmarks.

\begin{table}[htbp]
    \setlength{\belowcaptionskip}{6pt}
    \centering\footnotesize
    \caption{\textbf{Flappy Bird controls and alternative readouts.}
    Counts use ten level seeds per setting. ``Hidden'' denotes predictor
    hidden states, and ``latent'' denotes predicted latent states.
    The last two controls use a two-frame CNN encoder, residual-MLP
    dynamics, and a 24-step consistency loss.}
    \label{tab:flappy-controls}
    \begin{tabular}{llrr}
        \toprule
        Model & Readout & Games clearing $\geq1$ pipe & Best pipes \\
        \midrule
        \method, likelihood only & Hidden & $10/10$ & 100 \\
        JEPA-style, invertible & Hidden & $1/10$ & 1 \\
        JEPA-style, invertible & Latent & $3/10$ & 1 \\
        JEPA-style, ViT & Hidden & $0/10$ & 0 \\
        JEPA-style, ViT & Latent & $4/10$ & 2 \\
        CNN + residual MLP & Latent & $0/10$ & 0 \\
        CNN + residual MLP + SIGReg & Latent & $0/10$ & 0 \\
        \bottomrule
    \end{tabular}
\end{table}

\FloatBarrier
\Needspace{8\baselineskip}
Changing the readout to predicted latents improves the number of games
clearing the first pipe for both JEPA-style controls, but their best
scores remain one and two pipes (Table~\ref{tab:flappy-controls}). In one
collision example, readouts on the real frame immediately before impact
predict death probabilities of 0.99 for \method and 1.00 for both controls.
After six imagined steps under the executed actions, the corresponding
probabilities are 0.97, 0.07, and 0.05. In this example, the deterministic
rollouts lose collision information that the readouts can recover from
real observations.

%% file: figures/decoded_rollouts.tex
\begingroup
\newcommand{\rolloutpanel}[2]{%
  \begin{minipage}[t]{0.485\linewidth}
    \centering\sffamily\small\textbf{#1}\par\vspace{3pt}
    #2%
  \end{minipage}%
}
\newcommand{\rolloutframe}[2]{%
  \begin{minipage}[t]{0.192\linewidth}
    \centering
    #1%
    #2%
  \end{minipage}%
}
{\small\bfseries (a) Teleport}\par\vspace{6pt}
\rolloutpanel{LeWM}{%
\rolloutframe{{\fontsize{7}{8}\selectfont $t+5$\par}\vspace{2pt}}{\includegraphics[width=\linewidth]{figures/decoded_frames/teleport/LeWM/t_plus_05.png}}\hfill
\rolloutframe{{\fontsize{7}{8}\selectfont $t+10$\par}\vspace{2pt}}{\includegraphics[width=\linewidth]{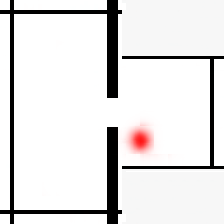}}\hfill
\rolloutframe{{\fontsize{7}{8}\selectfont $t+15$\par}\vspace{2pt}}{\includegraphics[width=\linewidth]{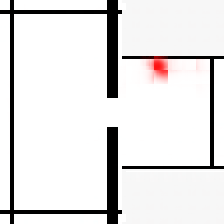}}\hfill
\rolloutframe{{\fontsize{7}{8}\selectfont $t+20$\par}\vspace{2pt}}{\includegraphics[width=\linewidth]{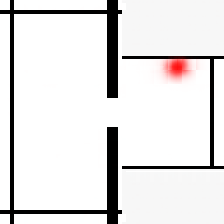}}\hfill
\rolloutframe{{\fontsize{7}{8}\selectfont $t+25$\par}\vspace{2pt}}{\includegraphics[width=\linewidth]{figures/decoded_frames/teleport/LeWM/t_plus_25.png}}%
}\hfill
\rolloutpanel{NFWM (upper)}{%
\rolloutframe{{\fontsize{7}{8}\selectfont $t+5$\par}\vspace{2pt}}{\includegraphics[width=\linewidth]{figures/decoded_frames/teleport/NFWM_upper/t_plus_05.png}}\hfill
\rolloutframe{{\fontsize{7}{8}\selectfont $t+10$\par}\vspace{2pt}}{\includegraphics[width=\linewidth]{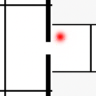}}\hfill
\rolloutframe{{\fontsize{7}{8}\selectfont $t+15$\par}\vspace{2pt}}{\includegraphics[width=\linewidth]{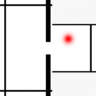}}\hfill
\rolloutframe{{\fontsize{7}{8}\selectfont $t+20$\par}\vspace{2pt}}{\includegraphics[width=\linewidth]{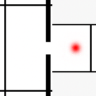}}\hfill
\rolloutframe{{\fontsize{7}{8}\selectfont $t+25$\par}\vspace{2pt}}{\includegraphics[width=\linewidth]{figures/decoded_frames/teleport/NFWM_upper/t_plus_25.png}}%
}
\par\vspace{7pt}
\rolloutpanel{SMWM}{%
\rolloutframe{{\fontsize{7}{8}\selectfont $t+5$\par}\vspace{2pt}}{\includegraphics[width=\linewidth]{figures/decoded_frames/teleport/SMWM/t_plus_05.png}}\hfill
\rolloutframe{{\fontsize{7}{8}\selectfont $t+10$\par}\vspace{2pt}}{\includegraphics[width=\linewidth]{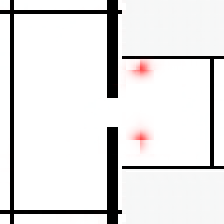}}\hfill
\rolloutframe{{\fontsize{7}{8}\selectfont $t+15$\par}\vspace{2pt}}{\includegraphics[width=\linewidth]{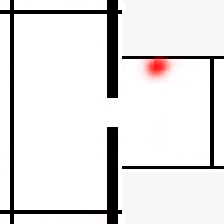}}\hfill
\rolloutframe{{\fontsize{7}{8}\selectfont $t+20$\par}\vspace{2pt}}{\includegraphics[width=\linewidth]{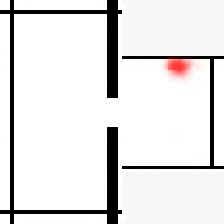}}\hfill
\rolloutframe{{\fontsize{7}{8}\selectfont $t+25$\par}\vspace{2pt}}{\includegraphics[width=\linewidth]{figures/decoded_frames/teleport/SMWM/t_plus_25.png}}%
}\hfill
\rolloutpanel{NFWM (lower)}{%
\rolloutframe{{\fontsize{7}{8}\selectfont $t+5$\par}\vspace{2pt}}{\includegraphics[width=\linewidth]{figures/decoded_frames/teleport/NFWM_lower/t_plus_05.png}}\hfill
\rolloutframe{{\fontsize{7}{8}\selectfont $t+10$\par}\vspace{2pt}}{\includegraphics[width=\linewidth]{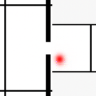}}\hfill
\rolloutframe{{\fontsize{7}{8}\selectfont $t+15$\par}\vspace{2pt}}{\includegraphics[width=\linewidth]{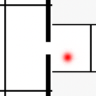}}\hfill
\rolloutframe{{\fontsize{7}{8}\selectfont $t+20$\par}\vspace{2pt}}{\includegraphics[width=\linewidth]{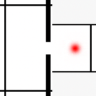}}\hfill
\rolloutframe{{\fontsize{7}{8}\selectfont $t+25$\par}\vspace{2pt}}{\includegraphics[width=\linewidth]{figures/decoded_frames/teleport/NFWM_lower/t_plus_25.png}}%
}
\par\vspace{7pt}
\vspace{3pt}
{\small\bfseries (b) Stochastic Boxing}\par\vspace{6pt}
\rolloutpanel{LeWM}{%
\rolloutframe{}{\includegraphics[width=\linewidth]{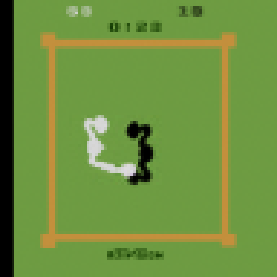}}\hfill
\rolloutframe{}{\includegraphics[width=\linewidth]{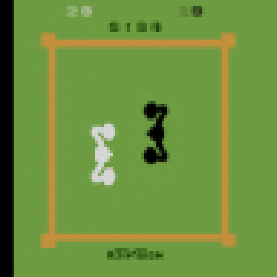}}\hfill
\rolloutframe{}{\includegraphics[width=\linewidth]{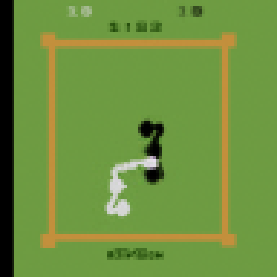}}\hfill
\rolloutframe{}{\includegraphics[width=\linewidth]{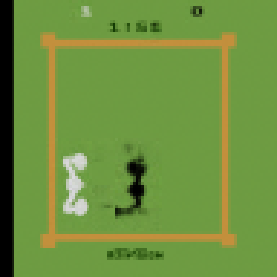}}\hfill
\rolloutframe{}{\includegraphics[width=\linewidth]{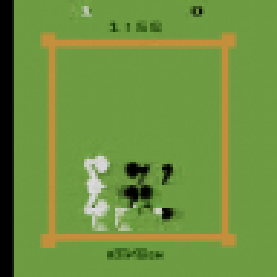}}%
}\hfill
\rolloutpanel{NFWM}{%
\rolloutframe{}{\includegraphics[width=\linewidth]{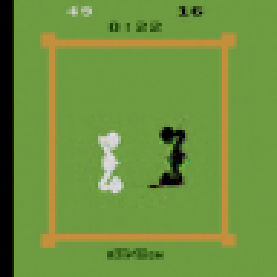}}\hfill
\rolloutframe{}{\includegraphics[width=\linewidth]{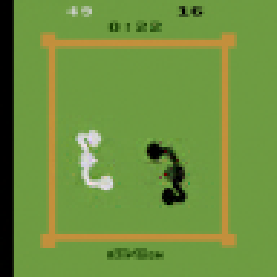}}\hfill
\rolloutframe{}{\includegraphics[width=\linewidth]{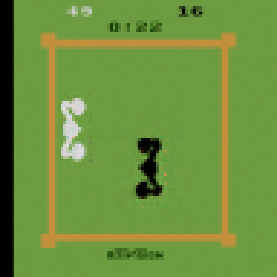}}\hfill
\rolloutframe{}{\includegraphics[width=\linewidth]{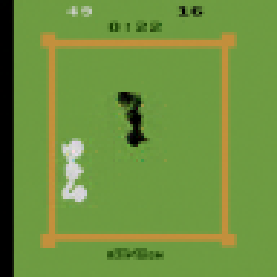}}\hfill
\rolloutframe{}{\includegraphics[width=\linewidth]{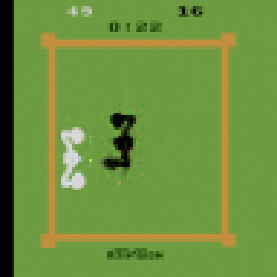}}%
}
\par\vspace{7pt}
\rolloutpanel{SMWM}{%
\rolloutframe{}{\includegraphics[width=\linewidth]{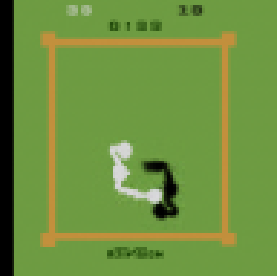}}\hfill
\rolloutframe{}{\includegraphics[width=\linewidth]{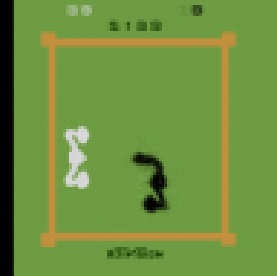}}\hfill
\rolloutframe{}{\includegraphics[width=\linewidth]{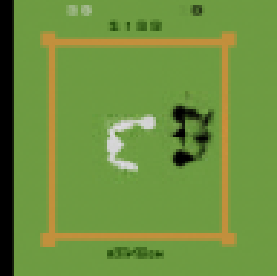}}\hfill
\rolloutframe{}{\includegraphics[width=\linewidth]{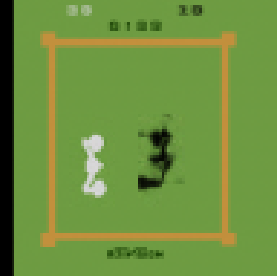}}\hfill
\rolloutframe{}{\includegraphics[width=\linewidth]{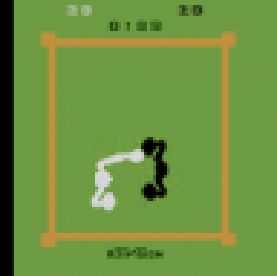}}%
}\hfill
\rolloutpanel{NFWM}{%
\rolloutframe{}{\includegraphics[width=\linewidth]{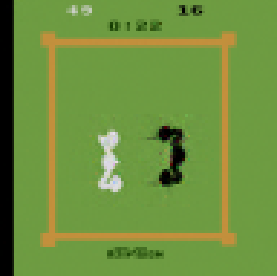}}\hfill
\rolloutframe{}{\includegraphics[width=\linewidth]{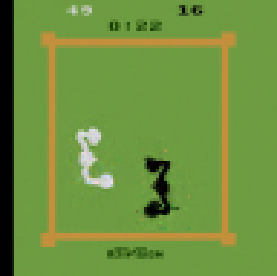}}\hfill
\rolloutframe{}{\includegraphics[width=\linewidth]{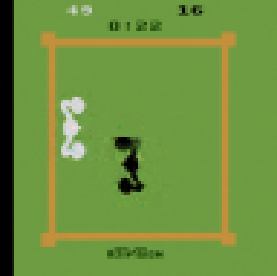}}\hfill
\rolloutframe{}{\includegraphics[width=\linewidth]{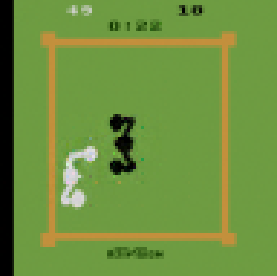}}\hfill
\rolloutframe{}{\includegraphics[width=\linewidth]{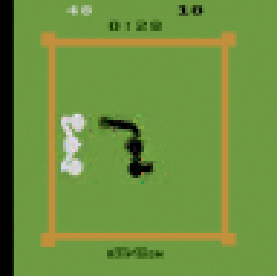}}%
}
\par\vspace{7pt}
\vspace{3pt}
\endgroup

%% file: main/appendix_followup.tex
\section{Additional Comparisons and Computational Cost}
\label{app:additional-comparisons}

\subsection{Full-Patch Predictive States}
\label{app:patch-state-followup}

The compact LeWM and SMWM baselines predict a 192-dimensional frame embedding.
For the full-patch controls, we match \method's $96\times96$ input,
$8\times8$-pixel patches, and $144\times192$ predictive state. We retain
all patch features and apply LeWM's SIGReg or SMWM's inverse-dynamics
objective to their concatenated representations. The controls use the
same task data and planning budgets as \method, with the optimization
settings in Table~\ref{tab:shared-model-settings}. We report epoch 10 on
Push-T and TwoRoom and epoch 30 on LIBERO, using one training seed per
full-patch model.

\begin{table}[htbp]
    \centering\footnotesize
    \caption{\textbf{Full-patch planning comparisons.} Success (\%). \method reports the main three-run mean. Each full-patch baseline uses one training run.}
    \label{tab:patch-state-followup}
    \begin{tabular}{lrrr}
        \toprule
        Task & \method & Full-patch LeWM & Full-patch SMWM \\
        \midrule
        Push-T & 89.3 & 80.0 & 28.0 \\
        TwoRoom & 98.0 & 40.0 & 98.0 \\
        LIBERO & 46.0 & 30.0 & 34.0 \\
        \bottomrule
    \end{tabular}
\end{table}

Table~\ref{tab:patch-state-followup} shows that retaining more coordinates does not consistently improve planning.
Full-patch SMWM matches \method on TwoRoom but performs substantially worse
on Push-T. On LIBERO, \method retains its advantage when both baselines
predict full patch states. Appendix~\ref{app:compute-followup} reports the
computational cost of these larger states.

\subsection{Rollout Counts and Terminal Scoring}
\label{app:rollout-budget-followup}

The primary planner uses one rollout per candidate and Final Mean scoring.
We vary the number $K$ of rollouts per candidate and compare two scores:
\begin{align*}
    C_{\mathrm{sample}}(a)
        &=\frac{1}{Kd}\sum_{k=1}^{K}\|\hat z^{(k)}_{t+H}-z_g\|_2^2,\\
    C_{\mathrm{mean}}(a)
        &=\frac{1}{Kd}\sum_{k=1}^{K}\|\mu^{(k)}_{t+H}-z_g\|_2^2.
\end{align*}
Earlier transitions are sampled under both rules. Each terminal mean
conditions on its rollout's sampled history. For a fixed history, the
expected sampled-endpoint score is
\[
    \mathbb E\!\left[d^{-1}\|\hat z_{t+H}-z_g\|_2^2\right]
    =d^{-1}\|\mu_{t+H}-z_g\|_2^2+d^{-1}\operatorname{tr}\Sigma_{t+H}.
\]
Final Mean omits the terminal variance term while retaining randomness
in earlier states.

\begin{table}[htbp]
    \centering\footnotesize
    \caption{\textbf{Success under different rollout scores (\%).} Each entry reports sampled endpoint / Final Mean, using one training run and 50 evaluation cases per task.}
    \label{tab:rollout-budget-followup}
    \begin{tabular}{lrrrr}
        \toprule
        Task & $K=1$ & $K=4$ & $K=8$ & $K=16$ \\
        \midrule
        Teleport & 44 / 42 & 30 / 26 & 30 / 30 & 30 / 26 \\
        Push-T & 92 / 92 & 90 / 92 & 92 / 92 & 92 / 92 \\
        LIBERO & 50 / 48 & 54 / 56 & 54 / 52 & 52 / 54 \\
        \bottomrule
    \end{tabular}
\end{table}

We use training seed 3072 and evaluation seed 100045 for Push-T and
Teleport, with epoch-10 and epoch-150 checkpoints, respectively. LIBERO
uses training seed 2, evaluation seed 3072, and the epoch-30 checkpoint.
All goals are encoded without noise. Push-T and Teleport use the CEM
budget in Table~\ref{tab:planning-settings}: 300 candidates, 30 elites,
30 iterations, and five predicted blocks of five actions. LIBERO uses
the same 50-candidate pools as the main ranking experiment. The task
cases and candidate budgets remain fixed across $K$. CEM search is run
for each scoring rule, so its later candidate populations can differ.
Increasing $K$ increases rollout compute. These single-run measurements
are reported separately from the three-run main results.

Neither score consistently yields higher success in Table~\ref{tab:rollout-budget-followup}. Push-T remains between
90\% and 92\%, LIBERO varies without a monotonic trend, and Teleport
success decreases with additional rollouts. Averaging more samples
estimates expected latent distance more accurately, but this distance
can favor a different action from the one with the highest success
probability. We retain Final Mean with $K=1$ as the default, avoiding
sampling noise in the final transition at low rollout cost.

\Needspace{8\baselineskip}
\subsection{Action Experts across LIBERO Suites}
\label{app:action-expert-followup}

Following the frozen-encoder protocol of TC-LeWM~\citep{liu2026tclewm},
we pretrain one \method on demonstrations from the Spatial, Object, Goal,
and Long suites, then train one action expert per suite. The 40 tasks
provide 2,000 demonstrations. We split each task into 45 training and
five validation demonstrations with seed 3072.

The model uses external and wrist-camera observations, resized to
$96\times96$. A shared invertible encoder processes each view into
$144\times192$ patch features. The six-layer predictor operates jointly
on the resulting 288 tokens, with three-frame context and five native
actions per transition. We train for 30,000 updates with batch size 128,
learning rate $5\times10^{-5}$, input-noise standard deviation 0.08, and
inverse-dynamics weight 10. The visual encoder is then frozen.

Each action expert uses a four-layer, width-384 transformer with six
attention heads and cross-attention to all 288 visual tokens. Its other
inputs are a learned task-ID embedding, flow time, and eight-dimensional
proprioception (end-effector position, axis-angle rotation, and gripper
joint positions). Proprioception enters only the action expert. Flow
matching trains the expert to predict eight seven-dimensional actions,
using a Gaussian source, uniform flow time, and a velocity-MSE loss.
We train each expert for 40,000 updates with batch size 128, learning
rate $2\times10^{-4}$, and 1,000 warm-up updates. All stages use seed 3072
and bfloat16 training. The policy uses clean images, ten Euler steps,
and executes each eight-action chunk before observing again.

We evaluate the final expert checkpoint on 50 initial states per task,
using evaluation seed 4242 and the native LIBERO success criterion.
Episodes allow 220, 280, 300, and 520 control steps for Spatial, Object,
Goal, and Long, respectively, after ten settling steps. Image rendering
matches the demonstrations. Successful trials are 369/500, 481/500,
450/500, and 354/500 across the four suites, yielding 82.7\% overall
success (Table~\ref{tab:action-expert-followup}).

\subsection{Latency, Rollout Throughput, and Decision Time}
\label{app:compute-followup}

We profile on an NVIDIA RTX PRO 6000 Blackwell Server Edition GPU.
Table~\ref{tab:runtime-followup} separates one-step latent prediction,
batched rollout, and complete decisions. Prediction uses batch size one.
Rollout throughput uses batches of 32 over five predicted steps. Peak
allocated rollout memory includes resident models, inputs, and
intermediate tensors. Push-T, TwoRoom, and LIBERO use float32 inference
for \method, LeWM, and SMWM. Boxing and all JEDI timings use bfloat16
mixed precision.

Decision timing covers CEM search on Push-T and TwoRoom, ranking the
50 candidates on LIBERO, and observation encoding plus policy execution
on Boxing. CEM uses the settings in Table~\ref{tab:planning-settings},
including five predicted action blocks. JEDI uses three diffusion steps.
Its Push-T and LIBERO timings use task-trained versions of the CNN
adaptation in Appendix~\ref{app:jedi}, with four-frame context and no
reward or termination heads.
DreamerV3 rebuilds its state from the last three frames at each decision.
Decision timers include encoding and model computation, with environment
execution and host-to-device data loading excluded. CUDA synchronization
delimits the timed workload. After three warm-up runs, prediction and
rollout measurements average ten and five repetitions, respectively.
Search and ranking average three repetitions, and Boxing policy decisions
average twenty. The DreamerV3 measurements average ten
repetitions for each operation after three warm-up runs.

\begin{table}[htbp]
    \centering\scriptsize
    \setlength{\tabcolsep}{4pt}
    \caption{\textbf{Prediction and decision cost.} Prediction latency (ms), rollout states/s, peak allocated rollout memory (MiB), and decision time (s).}
    \label{tab:runtime-followup}
    \begin{tabular}{llrrrr}
        \toprule
        Task & Model & Predict (ms) & States/s & Peak (MiB) & Decision (s) \\
        \midrule
        Push-T & \method & 1.19 & 3,586 & 423.5 & 10.038 \\
        & LeWM & 1.17 & 28,491 & 84.2 & 0.286 \\
        & SMWM & 1.17 & 28,431 & 83.2 & 0.286 \\
        & LeWM, full patch & 1.18 & 3,521 & 430.0 & 10.478 \\
        & SMWM, full patch & 1.20 & 3,533 & 484.0 & 10.478 \\
        & JEDI & 5.16 & 6,230 & 61.9 & 1.416 \\
        \midrule
        TwoRoom & \method & 1.22 & 3,568 & 423.5 & 10.080 \\
        & LeWM, full patch & 1.20 & 3,411 & 430.0 & 10.499 \\
        & SMWM, full patch & 1.20 & 3,522 & 484.0 & 10.517 \\
        \midrule
        LIBERO & \method & 1.21 & 3,710 & 423.6 & 1.655 \\
        & LeWM & 1.21 & 29,155 & 84.3 & 1.593 \\
        & SMWM & 1.17 & 29,014 & 84.6 & 1.589 \\
        & LeWM, full patch & 1.18 & 3,416 & 430.0 & 1.633 \\
        & SMWM, full patch & 1.28 & 3,463 & 484.3 & 1.643 \\
        & JEDI & 5.17 & 6,197 & 61.9 & 6.576 \\
        \midrule
        Boxing & \method & 1.54 & 7,501 & 281.9 & 0.00326 \\
        & LeWM & 1.42 & 22,119 & 82.0 & 0.00304 \\
        & SMWM & 1.41 & 22,038 & 82.4 & 0.00304 \\
        & JEDI & 5.12 & 6,235 & 83.8 & 0.00127 \\
        & DreamerV3 (3-frame memory) & 2.06 & 15,425 & 98.6 & 0.00322 \\
        \bottomrule
    \end{tabular}
\end{table}

\begin{samepage}
On Push-T, \method has similar planning cost to the full-patch controls
and higher cost than the compact baselines. In Boxing, it has lower
single-step latency than JEDI and DreamerV3, while DreamerV3 has higher
rollout throughput and JEDI has lower policy-inference latency.
Predictive-state size and the downstream workload both affect runtime.
\par
\end{samepage}

\paragraph{Parallel and token-wise Teleport prediction}
Table~\ref{tab:teleport-compute} uses the same CEM budget for all models:
300 candidates, 30 elites, 30 iterations, and five blocks of five actions.
Each timing records one complete decision after warm-up.
Both \method predictors sample all transitions, including the endpoint.
The planner starts from one observation and a clean goal. Timing covers
context and goal encoding, search, and scoring, excluding environment
execution. A decision takes 22.3 times as long with token-wise prediction as with parallel prediction,
with peak allocated memory increasing from 3.48 to 9.82\,GiB.

\begin{table}[htbp]
    \centering\footnotesize
    \caption{\textbf{Teleport planning cost.} Peak allocated memory covers the complete CEM decision.}
    \label{tab:teleport-compute}
    \begin{tabular}{lrr}
        \toprule
        Model & Decision (s) & Peak memory (GiB) \\
        \midrule
        Parallel \method & 10.14 & 3.48 \\
        Token-wise \method & 225.92 & 9.82 \\
        LeWM & 0.34 & 0.11 \\
        SMWM & 0.38 & 0.10 \\
        \bottomrule
    \end{tabular}
\end{table}